%% file: acl_latex.tex
\documentclass[11pt]{article}

\usepackage[final]{acl}

\usepackage{times}
\usepackage{latexsym}
\usepackage{graphicx}
\usepackage{colortbl}
\usepackage{multirow}
\usepackage{booktabs}
\usepackage{subcaption}
\usepackage[export]{adjustbox}
\usepackage{newunicodechar}
\newunicodechar{Ṭ}{T}

\usepackage{algorithm}
\usepackage{algpseudocode}
\usepackage{float}
\usepackage{hyperref}
\usepackage[]{acronym} 

\usepackage{comment}
\usepackage{tabularx}
\usepackage[T1]{fontenc}

\usepackage[utf8]{inputenc}
\usepackage[T1]{fontenc}
\usepackage{microtype}

\usepackage{inconsolata}
\usepackage{makecell} 
\usepackage{devanagari}

\usepackage{amsmath}

\title{Digitizing Nepal's Written Heritage: \\ A Comprehensive HTR Pipeline for Old Nepali Manuscripts}

\author{
  Anjali Sarawgi$^{1*}$~~~
  Esteban Garces Arias$^{1,3}$~~~
  Christof Zotter$^{2}$\\
  $^1$Department of Statistics, LMU Munich, 
  $^2$Heidelberg Academy of Sciences and Humanities\\
  $^3$Munich Center for Machine Learning (MCML)\\[0.5em]
  $^*$Corresponding author: \url{Anjali.Sarawgi@campus.lmu.de}
}

\begin{document}
\maketitle



\begin{abstract} 
This paper presents the first end-to-end pipeline for Handwritten Text Recognition (HTR) for Old Nepali, a historically significant but low-resource language. We adopt a line-level transcription approach and systematically explore encoder-decoder architectures and data-centric techniques to improve recognition accuracy. Our best model achieves a Character Error Rate (CER) of 4.9\%. In addition, we implement and evaluate decoding strategies and analyze token-level confusions to better understand model behavior and error patterns. Although the evaluation dataset is confidential, we release our training code, model configurations, and evaluation scripts to support further research on HTR for low-resource historical scripts.

\end{abstract}


\section{Introduction}

Handwritten Text Recognition (HTR) remains a challenging task—particularly for historical and low-resource languages such as Old Nepali. Historical documents pose numerous challenges for offline handwriting recognition systems, especially during the segmentation and labeling stages \cite{10.1007/978-3-030-01231-1_23, 8395169}. Nepal, particularly the Kathmandu Valley, has a rich manuscript culture that features diverse scripts for various languages, including Newari (also referred to as Nepal Bhasha), Sanskrit, and Maithili. For the text recognition of Pracalit Lipi, the most common script in the Valley for centuries, we refer to \citet{nakarmi-etal-2024-nepal}. The historical manuscripts used in this study are written in Old Nepali, the language of the Gorkhalis who conquered the Kathmandu Valley kingdoms in the late 18th century. The new government used the Devanagari script, which is still dominant in modern Nepal and many parts of India. 

\begin{figure}[H]
    \centering
    \setlength{\fboxsep}{0pt} 
    \setlength{\fboxrule}{1.5pt} 
    \fbox{\includegraphics[width=0.4\textwidth]{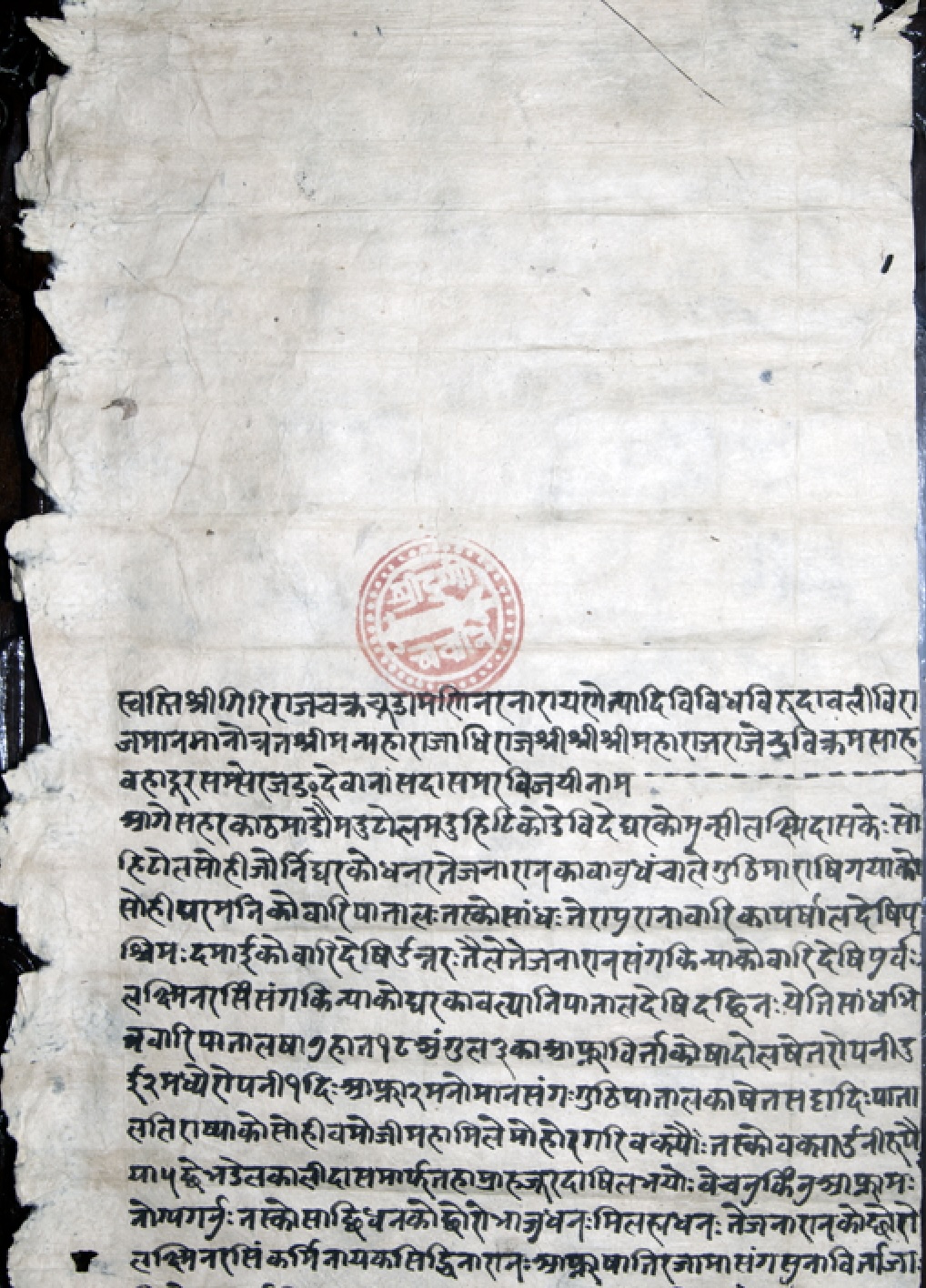}}
    \caption{Sample manuscript containing Old Nepali in Devanagari script. This image is sourced and cropped from the Documenta Nepalica~\cite{saha1832documenta}, courtesy of Manik Bajracharya.}
    \label{fig:Old_Nepali_Manuscript}
\end{figure}

However, the historical manuscripts considered in this study present several challenges, including diverse handwriting styles, degraded document quality, intricate conjunct forms, and limited annotated data \cite{10.1108/JD-09-2023-0183}. These constraints necessitate approaches that leverage transfer learning, data augmentation, and carefully designed model architectures \citep{arias-etal-2023-automatic}. In this paper, we address these challenges by conducting a comprehensive exploration of modern HTR techniques for Old Nepali manuscripts. We investigate transfer learning strategies to enable effective learning in low-resource settings, implementing a three-stage approach that adapts models from high-resource settings to target-domain data. Our architectural exploration includes transformer-based approaches such as TrOCR \citep{li2023trocr}, as well as combinations of vision encoders (including Swin Transformer \citep{Liu2021Swin}) with language decoders (BERT-style \citep{Devlin2018BERT} and GPT-2 \citep{Radford2019GPT2}). To effectively capture the intricacies of Old Nepali scripts, we compare byte-level tokenization with traditional approaches to determine the optimal granularity for this complex writing system. We further enhance model robustness through data augmentation strategies that preserve script integrity while improving generalization \cite{rassul2025advancingofflinehandwrittentext}. Finally, we systematically compare decoding methods—including deterministic methods (greedy, beam search \citep{Freitag_2017}, contrastive search \citep{su2022contrastive}) and stochastic methods (sampling with temperature \citep{ackley1985learning}, top-$k$ \citep{fan2018hierarchical}, and top-$p$ sampling \citep{holtzman2019curious})—to optimize recognition accuracy.\\
\noindent \textbf{Contributions:} (i)  We develop a comprehensive HTR pipeline that efficiently performs recognition on Old Nepali texts, demonstrating superior performance compared to both open-source and proprietary alternatives in this low-resource setting. (ii) We present a detailed analysis of preprocessing and data augmentation methods specifically tailored for historical scripts. (iii) We demonstrate the impact of domain adaptation through pretraining on datasets derived from publicly available sources, including the Heidelberg open research data \citep{DATA/EGOKEI_2022}. (iv) We conduct a thorough investigation of tokenization and decoding methods to capture the unique structural properties of historical Nepali texts, revealing insights about optimal representation strategies for complex scripts. (v) We make our code, models, and analysis tools publicly available at \url{https://github.com/anjalisarawgi/nepOCR/}.

\section{Related Work}
HTR has evolved from traditional pattern recognition methods to deep learning approaches. Early systems were based on RNNs and the CTC framework developed by \citet{NIPS2008_66368270}, while later works combined CNNs with RNNs and CTC decoding \citep{Shi2017CRNN}. 

The advent of Tesseract OCR \citep{Smith2007Tesseract} provided a widely used baseline, albeit with limited accuracy for historical scripts. Recent advances in HTR have been driven by transformer-based architectures and deep learning approaches \cite{jimaging10010018}. TrOCR \cite{li2023trocr}, which leverages the Transformer architecture for both image understanding and wordpiece-level text generation, has shown significant improvements over traditional CNN-RNN approaches \cite{arias-etal-2023-automatic}. However, applying these state-of-the-art methods to low-resource historical scripts requires careful adaptation. Handwritten text recognition in low-resource scenarios, such as manuscripts with rare alphabets, is a challenging problem \cite{SOUIBGUI202243}. 

In addition to model architectures, recent research has focused on data augmentation for low-resource scripts \citep{Gupta2016SynthText} and the use of synthetic data to overcome the scarcity of annotated manuscripts. Prior studies \cite{Wigington2017, deSousaNeto2024} have also shown the value of data augmentation for HTR tasks. Tokenization methods, a critical component of natural language processing, have been extensively explored in the context of subword regularization \citep{Kudo2018Subword} and have shown promise for adapting language models to complex scripts. Recent studies have also examined the interplay between vision encoders and language decoders. For example, combinations involving Swin Transformers with BERT or GPT-2 have demonstrated promising results in text recognition tasks \citep{arias-etal-2023-automatic, koch-etal-2023-tailored, pavlopoulos-etal-2024-challenging}. Furthermore, novel decoding strategies such as contrastive search, adaptive contrastive search \citep{garces-arias-etal-2024-adaptive}, GUARD \cite{ding2025guardglocaluncertaintyawarerobust}, and Min-$k$ \cite{ding2026minksamplingdecouplingtruncation} have been proposed to overcome the limitations of beam search across various NLP tasks.


\section{Data Preparation}
\label{sec:data}


\subsection{Handwritten scripts}
\label{sec:cards}
The majority of the manuscripts analyzed in this study originate from the collection of the Nepal German Manuscript Preservation Project (NGMPP).\footnote{For further details, see Documenta Nepalica: \url{https://nepalica.hadw-bw.de/nepal/}.} The NGMPP undertook extensive microfilming of manuscripts and related documentary materials from a range of Nepalese repositories, including the National Archives in Kathmandu, as well as private collections. The selected examples date to the late eighteenth and nineteenth centuries, with many consisting of royal edicts bearing the red seal of the Shah kings. They typically have a consistent visual and structural layout (as illustrated in Figure~\ref{fig:Old_Nepali_Manuscript}) and follow the diplomatic conventions of the period. They begin with an invocation to a deity, followed by a blank space where the royal seal is imprinted. The main text begins with the standard Sanskrit panegyric of the king, names the addressee, discusses the subject matter of the document, and ends with the \textit{eschatocol} containing the date and a blessing \cite{Cubelic_2018, pant1989mustang}. The selection also includes well-written petitions to the king that share some diplomatic elements but typically have a simpler structure. Most of these documents consist of one or two pages, providing a limited number of training samples. To improve the dataset, we also included larger samples from two chronicle manuscripts that share a similar context and writing style.

As with South Asian manuscript cultures in general, \textit{scriptio continua} is the norm here. Only a few more recent documents include spaces to separate words. Hand-drawn bullets or other punctuation symbols, such as dandas, may appear at the end of a word, sentence, or clause. However, these symbols are used inconsistently, and bullets can also appear within words. As a result, detecting word and sentence boundaries remains challenging.

Devanagari is a rich, left-to-right abugida script and includes many complex conjunct characters (cf. Figures \ref{fig:sample_1} and \ref{fig:sample_2}). The form used in the documents differs slightly from present-day Devanagari. In addition, variations in handwriting, scribal styles, and even pen types introduce additional inconsistencies, making text recognition more challenging. Although the texts may include passages in other languages (such as the Sanskrit panegyrics), the main language is Old Nepali. Morphologically, this stage of language development is quite close to modern Nepali but differs, especially in orthography, which is highly variable \citep{riccardi1971}. Due to a lack of standardization, scribes often spelled the same word differently, even within a single document. Another challenge is the abundance of now-obsolete technical terms. The transcribed dataset used in this study consists of 155 manuscript images. The manuscripts are written in Devanagari, and an average page contains 1,198 characters across 20 lines. The lines are relatively long, averaging 60 characters. The average dimensions of the manuscript images are $3091 \times 3487$ pixels at 328 dpi, and a detailed character set is provided in Appendix~\ref{a:chars}.


\subsection{Image preprocessing}
\label{sec:preprocess}

We preprocess the Old Nepali dataset using line-level segmentation with Kraken's \citep{kiessling:hal-04936936} polygon-based segmentation method to detect and isolate individual text lines in the manuscript images. Each detected line is then cropped and used as an independent training sample, allowing the model to focus solely on text recognition. From a total of 155 manuscripts, we extracted 3,100 lines. The longest sequence contains 162 characters, and the shortest line consists of only 1 character. The standard deviation of line lengths is 27.65 characters, indicating high variation. The images are typically very wide and short, with an average size of $1593 \times 133$ pixels (width $\times$ height). An example of a cropped line image after preprocessing is shown in Figure~\ref{fig:preprocess_line_sample}, and detailed statistics are provided in Appendix~\ref{a:data_stats}.

\begin{figure}[h!]
    \centering
    \setlength{\fboxsep}{0pt} 
    \setlength{\fboxrule}{0.5pt} 
    \fbox{\includegraphics[width=\linewidth]{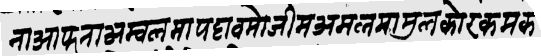}}
    \caption{Sample line image after preprocessing.}
    \label{fig:preprocess_line_sample}
\end{figure}


\subsection{Three-stage training dataset preparation}

Our training pipeline comprises three stages that adapt the model to Old Nepali manuscripts (see Appendix~\ref{a:training_pipeline}). This setup is important for two reasons: (1) the limited availability of labeled handwritten Devanagari data and (2) the need for a model that understands Devanagari script and its visual and stylistic variations. In the \textbf{first stage}, the model is trained on a large synthetic dataset to learn general Devanagari patterns. The \textbf{second stage} uses a printed Devanagari dataset to bridge the gap between synthetic and real data. Finally, in the \textbf{third stage}, the model is fine-tuned on the Old Nepali manuscript dataset. Here, the first and second stages function as pretraining to enable the model to learn general context and visual patterns, and the third stage performs final fine-tuning on the manuscript data.

\paragraph{First stage} We train on a synthetic dataset constructed using text extracted from historical Nepali textbooks available through the Internet Archive \cite{internetarchive}. The books from which the text corpus is extracted are listed in Table~\ref{tab:oldNepaliSynth_books}, in Appendix~\ref{a:fonts_and_noise}. These texts are rendered as line-level grayscale images using the Python Imaging Library (PIL), with a variety of Devanagari fonts. We generate 105,000 images and simulate script-level degradation by applying several noise and distortion effects. Specifically, we use a total of 11 fonts and 10 types of noise and distortion effects (e.g., geometric distortions, blurring, contrast or brightness variations), and the choice of font type and augmentation method for each image is randomized. A detailed overview is provided in Table~\ref{tab:fonts} and Table~\ref{tab:noise_synth} in Appendix~\ref{a:fonts_and_noise}. 

\begin{figure}[h!]
    \centering
    \setlength{\fboxsep}{0pt} 
    \setlength{\fboxrule}{0.5pt} 
    \fbox{\includegraphics[width=\linewidth]{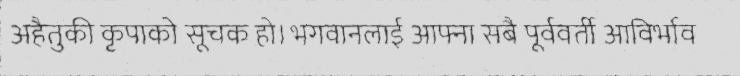}}
    \caption{Sample data for the first stage.}
    \label{fig:first_stage_sample}
\end{figure}

This synthetic dataset is prepared to mirror the real dataset. All images are at the line level, as we render individual lines of a textbook. From this synthetic corpus, we use 100,000 images for training and set aside 2,500 images each for validation and testing. This stage serves as an initial training phase for the decoder, helping it learn domain-relevant linguistic patterns. A sample from the synthetic dataset used in the first stage is shown in Figure~\ref{fig:first_stage_sample}.

\begin{figure}[h!]
    \centering
    \setlength{\fboxsep}{0pt} 
    \setlength{\fboxrule}{0.5pt} 
    \fbox{\includegraphics[width=\linewidth]{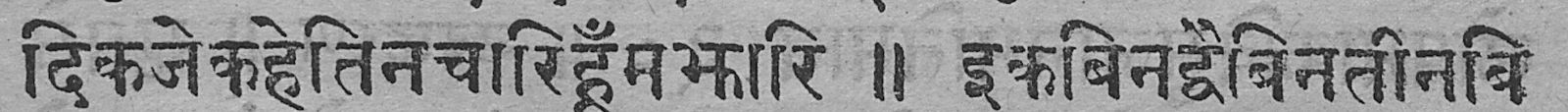}}
    \caption{Sample data for the second stage.}
    \label{fig:second_stage_sample}
\end{figure}

\paragraph{Second stage}  We train on a printed Devanagari dataset sourced from heiDATA \cite{DATA/EGOKEI_2022}, the open research data repository of Heidelberg University. As with the Old Nepali manuscripts, we apply preprocessing and line-level segmentation to scanned pages, yielding 5,139 line images from 247 full-page scans. The images are converted to grayscale to match the visual appearance of the Old Nepali manuscript data. We use 80\% of these line images for training, and allocate 10\% each for validation and test sets. While this dataset contains printed rather than handwritten script, it includes realistic noise from scanned script documents, making it useful as a transfer stage to bridge the gap between synthetic and handwritten data. A sample from the printed Devanagari dataset used in the second stage is shown in Figure~\ref{fig:second_stage_sample}.

\paragraph{Third stage}
We fine-tune on the target Old Nepali manuscript dataset of 3,100 line-level images from 155 historical manuscripts (80/10/10 train/validation/test split). This final stage adapts the model to the specific handwriting styles, orthographic variations, and degradation patterns unique to these historical documents. \noindent Table~\ref{tab:data_config} summarizes the data split across all training stages. For a fair comparison, we split the dataset once with a fixed seed and utilize the same partitions.

\begin{table}[h]
\centering
\resizebox{\columnwidth}{!}{%
\input{fig_tables/data_config}
}
\caption{Data configuration for the three-stage training.}
\label{tab:data_config}
\end{table}


\section{Experimental Setup}
\label{sec:exp}

This section outlines the model configurations and data refinement strategies used in our experiments. We first compare TrOCR encoder variants, language decoders, and tokenizer types. We then introduce a set of data-centric modifications designed to boost model performance in the low-resource setting of Old Nepali handwritten text. 

\subsection{Model selection}
We use two TrOCR variants, trocr-base-handwritten and trocr-large-handwritten, which are ViT-based encoders pretrained on printed and handwritten datasets. A Swin Transformer encoder is also evaluated as an alternative architecture. For decoders, we explore two types of language models, GPT-2 and BERT, and evaluate different tokenization methods to assess their effectiveness for Devanagari scripts. This is motivated by the need for Devanagari-aware decoding and by the question of whether script-specific tokenization improves performance. We focus on transformer-based models, as prior work, e.g., \citet{arias-etal-2023-automatic}, shows that they outperform CNN-RNN architectures on historical HTR tasks.

The combinations of encoder, decoder, and tokenizer result in a total of $3 \times 2 \times 2 = 12$ model configurations. Each configuration is trained using our three-stage setup: 6 epochs for pretraining, 10 for transfer learning, and 20 for final fine-tuning, totaling $12 \times 3 = 36$ runs. All stages use a fixed learning rate of \texttt{3e-5} with the AdamW optimizer, and a batch size of 8 (see Table~\ref{tab:params_tr} for more details). Table~\ref{tab:model_architecture_results} in Appendix~\ref{a:model_arch} presents the results for all model configurations. Model selection is based on validation performance at the final fine-tuning stage, where the model with the lowest Character Error Rate (CER) is selected.

\begin{table}[H]
    \centering
    \resizebox{\columnwidth}{!}{
    \begin{tabular}{lccc}
        \toprule
        \textbf{Parameter} & \textbf{First Stage} & \textbf{Second Stage} & \textbf{Third Stage} \\ \midrule
        Seed & 42 & 42 & 42 \\
        Optimizer & AdamW & AdamW & AdamW \\
        Epochs & 6 & 10 & 20 \\
        Batch size (Train/Val/Test) & 8 & 8 & 8 \\
        Learning rate & 3e-5 & 3e-5 & 3e-5 \\
        Weight decay & 0.01 & 0.01 & 0.01\\
        Warmup steps & 500 & 500 & 500\\
        \bottomrule
    \end{tabular}
    }
    
    \caption{Training hyperparameters for all three learning stages.}
    \label{tab:params_tr}
\end{table}

\subsection{Data-centric enhancements}
After selecting the best-performing model architecture, we conduct a series of controlled experiments to measure the impact of preprocessing and dataset refinement on the final model performance. Since our labeled dataset is very small, we focus on strategies that help the model maximize its learning on each training sample. Specifically, we apply data augmentations to introduce a wider range of visual variations during training, thereby expanding dataset diversity and improving model generalization. We also perform light label cleaning to reduce noise and inconsistencies in the ground truth that might confuse the model and distract it from learning meaningful character-level patterns.

\paragraph{Binarization}  
We evaluate the impact of binary thresholding as a preprocessing step for line images. Contrary to our hypothesis that binarization would improve model performance, the results indicate otherwise, with binarization slightly worsening performance by 1\% CER (see Appendix~\ref{a:binarization} for details). Thus, we decided to use non-binarized images for training.

\paragraph{Transcription normalization}  
To ensure consistency and stable training, we apply a series of stylistic, punctuation, and Unicode normalization steps to the transcriptions. This is particularly important in Old Nepali HTR, where even minor transcription inconsistencies can introduce ambiguity and degrade the model's performance. For example, characters such as \texttt{U+0310} and \texttt{U+0901} both represent the Devanagari chandrabindu but may appear inconsistently across samples, introducing noise and inconsistencies during training.  Given the limited size of our dataset, these inconsistencies introduce unnecessary variation that can hinder the model's learning. More details on the normalization steps are provided in Appendix~\ref{a:transcription_normalization_stats}.

\paragraph{Data Augmentation}  
To enhance model robustness against variability in handwriting, script style, and scan quality, and to expand the effective size of the training set through synthetic samples, we apply a wide range of image-level augmentations. These augmentations are moderate and realistic, aiming to mimic the distortions and degradations commonly found in handwritten historical scripts, without damaging the underlying script structure.

We group our augmentations into three categories. The first group includes \textit{shape and angle distortions}, such as small-angle rotations, axis shifts, horizontal and vertical stretching, and image tilts and skews, to simulate images acquired at oblique angles. These simulate camera angle, scanning inconsistencies, or physical wrapping of the document. The second group comprises quality-degradation augmentations, including Gaussian blur and noise, blurring in one direction, salt-and-pepper noise, JPEG compression artifacts, and jitter to introduce pixel-level noise. These transformations simulate common sources of noise arising from scanning hardware, paper texture, or manuscript age. The third group focuses on character-level distortions and is comparatively stronger than the other two groups. It includes smudging random parts of the text with small blurring boxes, elastic warping (to mirror crumpled pages), subtle text curvature (as from slightly bent documents), and mild morphological operations such as erosion, dilation, and edge sharpening. These help the model handle character-level degradations frequently observed in noisy or damaged manuscripts.

We use 20 augmentation variations and apply augmentation intensities of 2$\times$, 4$\times$, 8$\times$, 12$\times$, and 16$\times$. A detailed list of augmentation functions is provided in Appendix~\ref{a:augmentation}.


\subsection{Model Architecture}
\label{sec:models}

\paragraph{Tokenizer}

We train two types of subword tokenizers using the Hugging Face tokenizers library \citep{Moi_HuggingFace_s_Tokenizers_2023}: \texttt{CharBPETokenizer} and \texttt{ByteLevelBPETokenizer}. Both implement the Byte Pair Encoding (BPE) algorithm \cite{sennrich2016neuralmachinetranslationrare}, which iteratively merges frequent token pairs until a target vocabulary size is reached. For our experiments, we fix the vocabulary size to 500 to balance the trade-off between coverage of textual variation with token frequency in training, which is particularly beneficial in low-resource settings \citep{arias-etal-2023-automatic}. To train the tokenizers, we construct a combined corpus comprising lines from all three stages of our HTR pipeline.

\paragraph{Image encoder}
We adopt the trocr-base-handwritten model as our primary image encoder. It uses a 12-layer Vision Transformer (ViT), and processes images resized to 384$\times$384 pixels. The encoder splits images into patches, converts them to vectors, and processes them through Transformer layers to learn visual features for text recognition. Although this model was not explicitly trained on the Devanagari script, its pretraining on diverse handwritten and printed scripts enables it to generalize reasonably well to handwritten Devanagari forms, especially given the lack of a large labeled dataset of handwritten Devanagari. To test architectural variants, we also evaluate a Swin Transformer encoder (swin-base-patch4-window7-224-in22k), which underperformed due to limited handwritten Devanagari data (see Appendix~\ref{a:swin}). In our work, the TrOCR image encoder remained the better choice.

\paragraph{Text decoder}
We evaluate two decoder architectures for text generation: a BERT-based model (114.8M parameters) and a GPT-2-based model (114.6M parameters). Both models are trained from scratch on the synthetic dataset in the first stage (see Table~\ref{tab:data_config}). To better adapt to our data, we use custom script-aware tokenizers designed for Devanagari text. The decoders follow standard base configurations (12 layers, 768 hidden size, 12 attention heads), with a language modeling head added for token prediction. The only modification is the vocabulary size, which is set to match our custom tokenizer. They are also paired with the same visual encoder, which allows direct comparison of decoder performance for HTR.

\paragraph{Performance metrics}
We evaluate model performance using CER, weighted CER (CER(w)), and exact match accuracy (ACC), i.e., the percentage of predicted lines that exactly match the text. In this work, we use CER as the primary metric for all model design choices. The CER is computed as:
\begin{equation}
CER = \frac{S+D+I}{N} = \frac{S+D+I}{S+D+C}
\label{eq:cer}
\end{equation}
\noindent where \textit{S}, \textit{D}, \textit{I} denote substitutions, deletions, and insertions respectively, and \textit{C} denotes correct characters. Additionally, to account for the varying length of the labels, which range from 1 to 162, we utilize the weighted CER:
\begin{equation}
\textit{weighted CER} = \frac{\sum_{i=1}^n l_i \cdot CER_i}{\sum_{i=1}^n l_i}
\label{eq:cer-weighted}
\end{equation}

\noindent where $l_i$ is the number of characters of label $i$, and $CER_i$ is the CER for example $i, i=1, \ldots, n$. 

During evaluation, we remove zero-width Unicode characters (\texttt{\textbackslash u200B}, \texttt{\textbackslash u200C}, and \texttt{\textbackslash u200D}) from both predictions and ground truth labels before computing CER. These characters are used in Devanagari to control conjunct forms and are present in the training data. However, since they are often added for rendering, we exclude them to focus on core transcription accuracy. 

\paragraph{Decoding strategies}
We test five decoding methods: beam search, contrastive search, temperature sampling, top-$k$, and top-$p$ sampling. For each strategy, we evaluate a range of hyperparameters to study their impact on HTR performance. A detailed overview is provided in Appendix~\ref{a:decoding_results}.


\section{Results}
\label{sec:res}


\subsection{Model selection}
\label{sec:arch_comp}
Table~\ref{tab:model_architecture_results} in Appendix~\ref{a:model_arch} presents the results of our comparative experiments across different decoder architectures and tokenization schemes. Overall, the choice of decoder architecture has only a small effect on final model performance. Interestingly, even the tokenizer appears relatively robust to architectural changes. Among all configurations, the BERT decoder paired with the byte-level BPE achieves the best performance, reaching a CER of 0.082 after all three training stages. This suggests that architectural tuning yields only limited gains, with BERT and byte-BPE achieving the most favorable CER. 

\subsection{Performance gains through label cleaning and augmentation}
We evaluated the impact of label normalization and image-level augmentations on model performance. As shown in Table~\ref{tab:ablation}, label cleaning alone reduced CER from 0.089 to 0.084. The most substantial gains were achieved through data augmentation, which reduced CER to 0.056 with 8$\times$ augmentation. Increasing it to 12$\times$ and 16$\times$ did not yield further improvements, indicating a performance plateau. These results confirm the importance of both label quality and visual diversity for improved performance in low-resource OCR tasks. 

\begin{table}[H]
\centering
\small
\resizebox{\columnwidth}{!}{%
\begin{tabular}{lccccc}
\toprule
\textbf{Ablation Step} & \textbf{Samples}  & \textbf{CER} & \textbf{CER(w)} & \textbf{ACC} \\
\midrule
Original Dataset & 2,480  & 0.089 & 0.090 & 22.9\% \\
+ Normalization & 2,480  & 0.084  & 0.084 & 21.6\%\\
+ Augmentation (2$\times$) & 7,440 & 0.067 & 0.068 & 26.7\% \\
+ Augmentation (4$\times$) & 12,400 & 0.060 &  0.061  &27.1\%\\
+ \textbf{Augmentation (8$\times$)} & \textbf{22,320} & \textbf{0.056}  & 0.057 & 29.4\%\\
+ Augmentation (12$\times$) & 32,240 & 0.056 & 0.056  &30.0\% \\
+ Augmentation (16$\times$) & 42,160 & 0.056 & 0.056 &29.7\% \\
\bottomrule
\end{tabular}
}
\caption{Ablation study showing the impact of normalization and data augmentation on model performance. The best results are shown in \textbf{bold}.
}
\label{tab:ablation}
\end{table}

\paragraph{Evaluating impact of encoder variants}
We compared \texttt{trocr-base-handwritten} and \texttt{trocr-large-handwritten} encoders to assess the effect of model size. As shown in  Table~\ref{tab:trocr_variants}, the larger variant achieved improved results by reducing CER from 0.056 to 0.049 with a corresponding weighted CER of 0.048. 

\begin{table}[H]
    \centering
    \resizebox{\columnwidth}{!}{%
    \begin{tabular}{lccc}
         \toprule
        \textbf{Encoder Variant} & \textbf{CER} &  \textbf{CER(w)} & \textbf{ACC} \\ 
        \midrule
        trocr-base-handwritten &  0.056 & 0.057 & 29.4\%  \\
        trocr-large-handwritten & \textbf{0.049} & \textbf{0.048} & \textbf{33.5\%}  \\
        \bottomrule
    \end{tabular}
    }
    \caption{Performance comparison of TrOCR encoder variants (\texttt{base} and \texttt{large}) under identical training conditions. The best results are shown in \textbf{bold}.}

    \label{tab:trocr_variants}
\end{table}

\subsection{Evaluating decoding methods}
We observe that different decoding strategies and hyperparameter settings have minimal effect on the overall performance. For instance, all beam sizes (1, 5, 10, 20) yield an identical weighted CER of 0.0483. Contrastive search with varying $k$ and $\alpha$, as well as sampling-based methods with different $\tau$, $k$, and $p$ values, consistently result in CER $\approx 0.0488$. These negligible differences (Appendix~\ref{a:decoding_results}) suggest that the decoding strategy has minimal impact on the achieved HTR performance.

\subsection{Benchmarking against HTR \& OCR tools}
\label{sec:bench}
We benchmark our pipeline against two baselines: (1) a full TrOCR model (including both encoder and decoder), and (2) Google Cloud Vision OCR. Since TrOCR is not pretrained on Devanagari scripts, it produces non-Devanagari outputs. To ensure a fair comparison, we fine-tuned TrOCR on our dataset using the same training configuration as our approach. The fine-tuned TrOCR baseline achieves a CER of 0.096. Additionally, Google Cloud Vision OCR performs poorly on our dataset, failing to capture conjuncts, diacritics, and punctuation (see Appendix~\ref{a:benchmarking} for a detailed overview). Furthermore, the default TrOCR decoder, when fine-tuned on the normalized and 8x-augmented dataset, achieves a CER of 0.059. In comparison, our decoder achieves a slightly lower CER of 0.056, with notably fewer parameters and 1.4$\times$ faster evaluation, indicating better efficiency without compromising accuracy (see Appendix~\ref{a:decoder_bench}).


\subsection{Error analysis}
\label{sec:error}

To assess the strengths and weaknesses of our model, we conduct an error analysis focusing on token confusions and label lengths (see Appendix~\ref{a:error_analysis} for visualizations). We construct confusion matrices by comparing model predictions on the test set with the corresponding ground truth labels, thereby identifying frequently misclassified characters. Inspection of the resulting heatmap shows that many errors are structured rather than random, following systematic patterns of handwriting variation. For example, {\dn y} (ya) is often confused with {\dn p} (pa), and {\dn t} (ta) with {\dn n} (na). This is likely because these character pairs exhibit high visual similarity in handwritten form, with the degree of similarity varying across handwriting styles. This suggests that the model is not hallucinating but making plausible mistakes due to visual ambiguity. Such structured errors can also be leveraged for post-correction methods. Furthermore, we examined which individual characters the model makes the most errors on. This analysis revealed that a small subset of 10 characters (out of 80 total) accounts for over half of all errors, i.e., 55.9\% (Figure~\ref{fig:most_prob_characters} and Table~\ref{fig:most_prob_characters_table}). This highlights the specific characters the model consistently struggles with and can be prioritized in future work. The overall distribution of CER across individual lines in the test set is presented in Figure~\ref{fig:cer_histogram}. We observe a strongly skewed distribution, with most lines exhibiting low CERs in the range 0--0.05. This indicates that, for most samples, the model predictions are highly accurate and require minimal or no correction. Additionally, we analyze how model performance varies with line length (see Figure~\ref{fig:cer_vs_line_length}). The weighted CER remains relatively low and stable across short and medium-length lines. However, the error rates increase sharply for lines longer than 120 characters. This spike is likely due to both the increased difficulty of longer sequences and the limited number of such samples, with only 26 out of 3,100 total lines exceeding 120 characters. We further analyzed the error patterns for longer lines and found two common patterns: (i) missing tokens at multiple positions and (ii) short segments (3--7 characters) with consecutive incorrect predictions. To investigate this, we segmented these lines into two parts and re-evaluated them using our model. Here, we observed a substantial drop in errors; for example, in one case, errors decreased from 23 to 4, and in another, from 30 to 2. This finding suggests that the model has difficulty handling longer sequences, and splitting them into shorter parts can improve model performance. Finally, the script-aware decoder enables us to explore model uncertainty by comparing the relative probabilities of the top predictions. This helps us flag about 27\% of all errors, and over half of these can be recovered from the top-3 predictions. This shows potential for future work in post-correction and human-in-the-loop review (see Appendix~\ref{a:token_uncert_analysis}).

\subsection{Inspecting domain shift}
\begin{figure}[H]
    \centering
    \setlength{\fboxsep}{0pt} 
    \setlength{\fboxrule}{0.5pt} 
    \fbox{\includegraphics[width=0.45\textwidth]{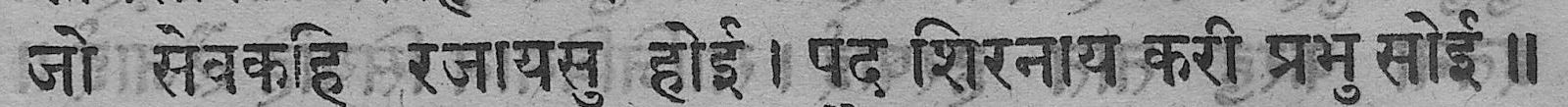}}
    \includegraphics[width=0.38\textwidth]{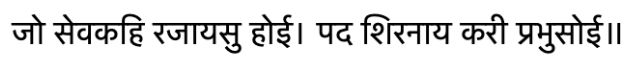}
    \caption{Printed Nagari script sample (top) and model's predicted transcription (bottom). The prediction is an exact match.}
    \label{fig:nagari}
\end{figure}

\begin{figure}[H]
    \centering
    \setlength{\fboxsep}{0pt} 
    \setlength{\fboxrule}{0.5pt} 
    \fbox{
    \includegraphics[width=0.45\textwidth]{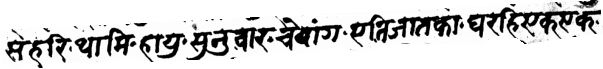}}
    \includegraphics[width=0.38\textwidth]{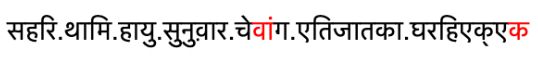}
    \caption{Old Nepali script sample (top) and model's predicted transcription (bottom). Mistakes are highlighted in red.}
    \label{fig:oldNepali}
\end{figure}

Pretraining on the printed Nagari script proved highly effective for our HTR task. Once fine-tuned, the model generalized well to handwritten data, despite the small size of the labeled dataset. This transfer learning approach was crucial for handling the low-resource setting. Figures~\ref{fig:nagari} and~\ref{fig:oldNepali} show the visual contrast between the printed and handwritten inputs the model was trained on, demonstrating how it adapted from clean printed forms to variable handwritten forms.
  

 We also evaluate the model at each training stage on the final test set to measure the contribution of each stage to the final model performance. The results in Table~\ref{tab:stage_effects} in Appendix~\ref{a:stage_training_effect} show that all three stages contribute positively to the model's final performance.

\section{Old Nepali Handwritten Text Recognition Application}
\label{sec:htr_app}
To demonstrate the practical applications of our approach, we have developed a web-based application that enables users to process images of Old Nepali manuscripts. This interactive tool allows researchers, historians, and linguists to upload photographs or scans containing Old Nepali handwritten text and receive automated analysis results.
The application performs two primary functions: (1) automatic line detection, which segments the manuscript into individual text lines, and (2) text recognition, which converts the handwritten content into machine-readable text. The system provides users with a visual overlay that displays the detected line boundaries alongside the recognized text, facilitating verification.
The application interface and sample processing results are illustrated in Appendix \ref{sec:app}. The web application is available at \url{https://nepocr.streamlit.app/}.

\section{Discussion and Outlook}
\label{sec:outlook}

Our experiments demonstrate advances in handwritten text recognition for low-resource historical scripts. The CER decreased by 49\%, from 9.6\% with the fine-tuned TrOCR baseline to 4.9\% with our final model. This improvement resulted from systematic enhancements to the HTR pipeline, as detailed in Figure \ref{fig:perf_gains} (Appendix \ref{sec:perf_gains}). The most substantial gains were achieved through three key innovations: script-aware decoder architecture, transcription normalization, and data augmentation combined with an enhanced vision encoder. These results confirm that modern transformer architectures can effectively adapt to non-Latin scripts when properly configured. Our three-stage training approach proved particularly valuable, with synthetic data bridging the gap between pretrained models and limited historical data. Interestingly, the choice of decoder architecture (BERT-style vs. GPT-2) and tokenization scheme had minimal impact on performance. This finding suggests that in low-resource settings, data quality and augmentation strategies matter more than model capacity. Indeed, our data-centric approaches—label normalization and data augmentation—produced the largest performance improvements. Error analysis provided additional insights into model behavior. Most character confusions follow predictable patterns based on visual similarity (e.g., {\dn y} vs. {\dn p}), indicating that the model learns meaningful visual representations rather than memorizing patterns. However, performance degrades on lines exceeding 120 characters, likely due to insufficient long-sequence examples in our training data. Several promising directions for future research emerge. First, collaborating with digital archives to expand the training corpus could help address current limitations in long-sequence data. Second, post-correction methods that exploit structured error patterns may further reduce CER, though this remains challenging because the dataset lacks explicit word boundaries. Third, applying our approach to other historical scripts would validate its generalizability.


\section{Conclusion}
\label{sec:concl}

We present a comprehensive pipeline for recognizing handwritten Old Nepali manuscripts, achieving a CER of 4.9\% through the systematic application of domain adaptation, data augmentation, and preprocessing. Our three-stage training approach helps bridge the gap between modern OCR models trained on Latin scripts and the specific challenges of historical Devanagari manuscripts, enabling effective learning in a low-resource setting. The experimental results suggest that data-centric improvements—particularly augmentation and label normalization—can contribute more to performance gains than architectural variations in this context, underscoring the importance of carefully curated training data. The model exhibits robust behavior across a broad range of handwriting styles and short-to-medium line lengths, with errors following interpretable patterns based on visual character similarity. These observations indicate that the model captures meaningful visual and structural features of the script, although some limitations remain, particularly for longer sequences and rare character combinations. This highlights the importance of expanding both the size and diversity of annotated datasets, as well as exploring complementary strategies such as post-correction or sequence-level modeling. Overall, this work provides a practical framework and a strong baseline for the digitization of historical manuscripts. While further validation on additional collections is necessary, extending the approach to other historical manuscript settings remains an important direction for future work. By releasing our code and models, we hope to support future research in historical document processing and contribute to ongoing efforts to preserve Nepal's written heritage.



\section*{Limitations}

While our approach demonstrates promising results, several limitations should be considered when applying this pipeline to the transcription of historical manuscripts. First, our study faces significant data constraints that limit the ability to assess generalizability. Although our data-centric approach partially mitigates this issue, our training corpus cannot fully capture the diversity of handwriting styles, document conditions, and formats found across historical archives. This limitation is particularly evident in the model's degraded performance on text lines exceeding 120 characters, which we attribute to the scarcity of long-sequence examples in our training data. 
Second, the limited public availability of training data poses a broader challenge to the field. Most datasets used in this study cannot be released publicly due to copyright restrictions, thereby hindering both accessibility and collaborative development of handwritten text recognition systems for historical documents. Nevertheless, we believe that releasing our trained model represents a valuable contribution for scholars working with Old Nepali manuscripts. Future work could also examine optimization parameters, such as warmup steps, which have been shown to be effective in recent studies \cite{Strbel2023}. Finally, our current pipeline relies on pre-segmented line inputs, which require external segmentation tools such as Kraken. This dependency introduces a potential source of error, particularly for documents with irregular layouts or severe physical damage. Future work should explore segmentation methods that adapt to the varied layouts commonly found in historical manuscripts.


\section*{Ethics Statement}

We affirm that our research adheres to the \href{https://www.aclweb.org/portal/content/acl-code-ethics}{ACL Ethics Policy}. This work does not involve human subjects or reveal any personally identifiable information. We declare that we have no conflicts of interest that could potentially influence the outcomes, interpretations, or conclusions of this research. All funding sources supporting this study are acknowledged in the acknowledgments section. We have made every effort to document our methodology, experiments, and results accurately and are committed to sharing our code and other relevant resources to foster reproducibility and further research advances.

\section*{Acknowledgments}

We would like to express our sincere gratitude to the colleagues in the research unit ``Documents on the History of Religion and Law of Pre-modern Nepal'' at the Heidelberg Academy of Sciences and Humanities for their assistance and for providing additional ground-truth data. We also thank Jan Kamlah of the University Library Mannheim for his constant support throughout this project. Finally, Esteban Garces Arias wishes to thank the Munich Center for Machine Learning (MCML) for their ongoing support.

\bibliography{custom}


\clearpage


\appendix

\begingroup
\let\clearpage\relax 
\onecolumn 
\endgroup

\section*{Appendix}

\section{Parameters for image binarization}
\label{a:hyper}
\begin{table}[H]
    \centering
    \begin{tabular}{l|l}
        \textbf{Method} & \textbf{Parameter} \\ \hline \hline
        Otsu's Binarization & Threshold = determined automatically by \texttt{cv2.THRESH\_OTSU} \\
    \end{tabular}
    \caption{Parameters for binarizing grayscale images \citep{4310076, opencv_library}.}
    \label{tab:binarization_param}
\end{table}

\section{Effect of Binarization on Model Performance}
\label{a:binarization}
\begin{table}[h]
\centering
\begin{tabular}{l|c|c}
\textbf{Method} & \textbf{CER} & \textbf{Used in training} \\
\midrule
\midrule
Binarized & 0.098 & No \\
Non-Binarized    & \textbf{0.089} & Yes \\
\end{tabular}
\caption{CER with and without binarization. Results are computed using three-stage training on the original Old Nepali dataset. Otsu's global thresholding is used for binarization \citep{4310076}. The best results are highlighted in \textbf{bold}.}
\label{tab:binarization_results}
\end{table}

\section{Model Architecture Comparison}
\label{a:model_arch}
\begin{table*}[ht]
\centering
\resizebox{1\textwidth}{!}{

\input{fig_tables/results_merged_valid}
}
\caption{Results for all model combinations after the 3-stage training. Model selection is performed based on the CER metric. The best results are highlighted in \textbf{bold}. All experiments were performed with an NVIDIA GeForce RTX 4090 GPU. We use the dataset with normalized transcriptions to reduce variance across runs and ensure a stable comparison. }
\label{tab:model_architecture_results}
\end{table*}

\section{Performance Metrics for Swin Encoder}
\label{a:swin}
\begin{table}[H]
    \centering
    \begin{tabular}{l|l|l}
        \textbf{Encoder variant} & \textbf{CER} & \textbf{ACC} \\ \hline \hline
        trocr-base-handwritten & \textbf{0.056} &  \textbf{29.35\%}  \\
        swin-base-patch4-window7-224-in22k & 0.174 & 21.93\% \\
    \end{tabular}
    \caption{CER and accuracy comparison between TrOCR and Swin encoders on the Old Nepali dataset. TrOCR outperforms Swin in both metrics for the same data and training configurations. The best results are highlighted in \textbf{bold}.}
    \label{tab:swin_encoder}
\end{table}

\section{Decoder Benchmarking}
\label{a:decoder_bench}
\begin{table}[H]
    \centering
    \begin{tabular}{l|l|l|l}
        \textbf{Decoder} & \textbf{CER} & \textbf{Total model parameters} & \textbf{Evaluation (samples/sec)}  \\ \hline \hline
        TrOCR decoder & 0.059 & 334M & 3.50 \\ 
        \textbf{Script-aware decoder (ours)} &  \textbf{0.056} & \textbf{202M} & \textbf{4.96} \\
    \end{tabular}
    \caption{Comparison of the standard TrOCR decoder (\texttt{trocr-base-handwritten}) with our script-aware decoder (with a BERT architecture, byteBPE tokenizer, vocab size = 500), under identical configurations using the normalized transcriptions and $8\times$ augmented dataset for fair comparison. The best results are highlighted in \textbf{bold}.}
    \label{tab:decoder_bench_table}
\end{table}

\section{Data Statistics}
\label{a:data_stats}
\begin{table}[H]
    \centering
        \begin{tabular}{l|l}
        \textbf{Statistic} & \textbf{Value} \\ \hline \hline
        
        \addlinespace[0.5ex]
        \multicolumn{2}{l}{\textit{Image-Level Statistics}} \\ \addlinespace[0.5ex] \hline
        Number of manuscripts & 155 \\
        Approximate time period & 18th and 19th centuries\\
        Mean characters per script &  1,198 \\
        Mean lines per script & 20 \\

        \addlinespace[1ex]
        \hline \addlinespace[0.5ex]
        \multicolumn{2}{l}{\textit{Line-Level Statistics}} \\ \addlinespace[0.5ex] \hline
        Total cropped lines & 3,100 \\
        Mean characters per line & 60 \\
        Shortest line length & 1 character \\
        Longest line length & 162 characters \\
        Standard deviation of line length & 27.65  \\
        Average image size (width × height) & 1593 × 133 pixels \\
    \end{tabular}
    \caption{Table showing detailed statistics of the dataset at full image (manuscript) and line level.}
    \label{tab:table_stats}
\end{table}

\begin{figure}[H]
    \centering
    \includegraphics[width=1.0\textwidth]{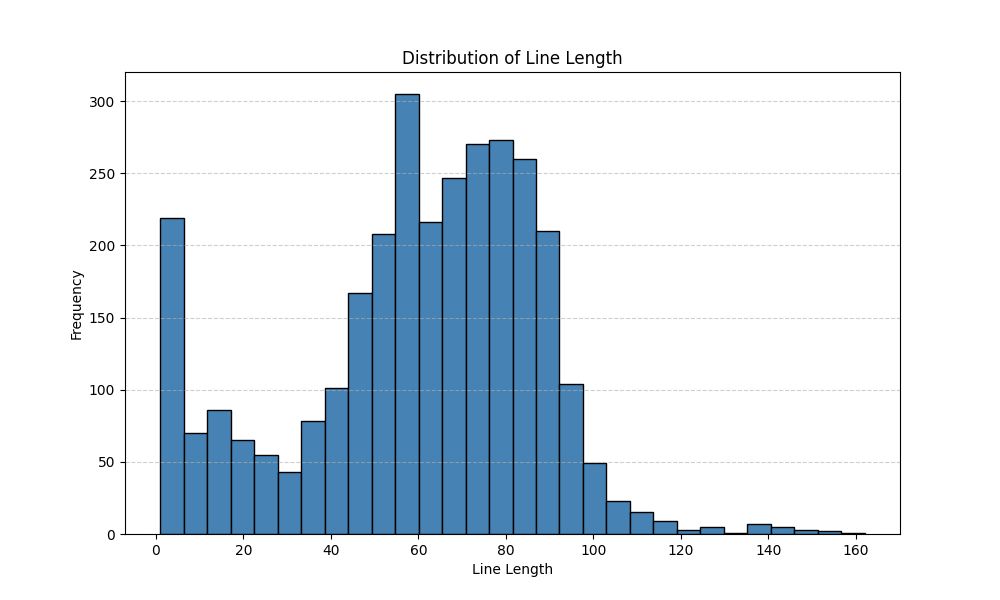}
    \caption{Histogram showing the distribution of the line lengths across all text samples in the dataset.}
    \label{fig:hist_line_length}
\end{figure}

\newpage

\section{Character Inventory}
\label{a:chars}

\begin{table}[H]
\centering
\resizebox{\textwidth}{!}{
\input{fig_tables/char_freq}
}
\caption{Character frequency and relative frequency observed in the ground truth labels.}
\label{tab:char_freq}
\end{table}

\clearpage

\section{Training Pipeline}
\label{a:training_pipeline}
\begin{figure}[H]
    \centering
    \includegraphics[width=0.5\textwidth]{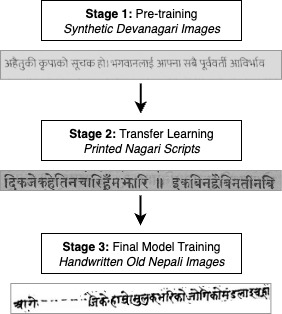}
    \caption{Visualization of the three-stage training pipeline with an example image from each dataset.}
    
\end{figure}

\section{Transcription Normalization Statistics}
\label{a:transcription_normalization_stats}

\begin{table}[H]
    \centering
    \begin{tabular}{l|l|l|l}
        \textbf{Normalization step} & \textbf{Count} & \textbf{Lines affected} & \textbf{\% of lines}  \\ \hline \hline
        Bullet symbols    & 4,965 & 1,780 & 57.41\% \\
        Total number of extra space fixed & 182 & 129 & 4.1\% \\
        Removing upper dashes (combining macrons) & 150 & 87 & 2.8\% \\
        Pipe to danda conversion & 3 & 3 & $<$0.1\%\\
        Convert ASCII digits to  Devanagari digits & 4  & 3 & $<$0.01\%\\
        Standardized chandrabindu & 3  & 3 & $<$0.1\%\\
    \end{tabular}
    \caption{Summary of character-level normalization steps applied during label cleaning. Each operation counts the number of characters that were modified or removed.}
    \label{tab:normalization}
\end{table}
\clearpage

\section{Augmentation Methods}
\label{a:augmentation}
\begin{table}[H]
    \centering
    \begin{tabularx}{\textwidth}{l|X}
        \textbf{Augmentation} & \textbf{Description} \\ \hline \hline
        \addlinespace[0.5ex]
        \multicolumn{2}{l}{ \textit{Shape and Angle Distortions}}\\ 
        \addlinespace[0.5ex] \hline
    
        \texttt{rotation} & Rotate the image by a small angle  ($\pm$3$^\circ$) \\
        \texttt{shift} & Shifts the whole image randomly by 5-20 pixels in the x and y directions \\
        \texttt{perspective} & Tilts the image to simulate a camera picture taken from an angle \\
        \texttt{shear} & Skews the image diagonally to mimic slanted text \\
        \texttt{hstretch} & Stretches the image horizontally by a factor of 1.2 \\
        \texttt{vstretch} & Stretches the image vertically by a factor of 1.2 \\
        \hline

        \addlinespace[0.5ex]
        \multicolumn{2}{l}{\textit{Noise and Blur Artifacts}} \\ \addlinespace[0.5ex] \hline
        \texttt{blur} & Applies Gaussian blur with kernel size 5$\times$ across the whole image  \\
        \texttt{motion\_blur} & Blurs in one direction to mimic movement during scanning \\
        \texttt{jpeg} & Simulates quality loss from JPEG compression \\
        \texttt{jitter} & Adds random pixel-level noise \\
        \texttt{gaussian\_noise} & Applies Gaussian noise to the full image, which adds random bright or dark speckles \\
        \texttt{elasticblur} & Combines elastic distortion with Gaussian blur \\
        \texttt{saltpepper} & Adds black and white specks like dust or scan noise  \\ \hline   

        \addlinespace[0.5ex]
        \multicolumn{2}{l}{\textit{Letter-Level Damage Simulation}}\\ \addlinespace[0.5ex] \hline
        \texttt{blurredpatches} & Picks random spots on the image and blurs to simulate smudges \\
        \texttt{sine} & Applies a wave horizontal distortion using sine curves \\
        \texttt{horizontal} & Curves the text horizontally, like bending the page gently \\
        \texttt{elastic} & Bends and wraps the image using smooth distortion, like crumpled paper  \\
        \texttt{med\_blur} & Applies a light median filtering with kernel size 3$\times$3, to smoothen very small irregularities  \\
        \texttt{morph} & Applies erosion or dilation to simulate ink spread or fade  \\
        \texttt{sharpen} & Enhances edge contrast using a 3$\times$3 convolutional kernel to simulate bold ink or high-contrast scanning  \\
    \end{tabularx}
    \caption{List of image-level augmentations used for generating synthetic data of Old Nepali image samples. All the augmentations are applied to grayscale, non-binarized images to preserve original stroke quality and noise characteristics.}
    \label{tab:aug_methods_main}
\end{table}


\clearpage
\section{Synthetic Dataset Preparation Fonts and Noise}
\label{a:fonts_and_noise}

\textbf{Fonts used for synthetic data generation}
\begin{table}[H]
    \centering
    \begin{tabular}{l|l}
        \textbf{Fonts} & \textbf{Source} \\ \hline \hline
        Akasha Regular & \href{https://fontesk.com/akasha-font/}{Link} \\
        Anek Devanagari & \href{https://fonts.google.com}{Google Fonts}  \\
        Chandas & \href{https://www.typingguru.net/}{Link}  \\
        IBM Plex Sans Devanagari & \href{https://fonts.google.com}{Google Fonts}  \\
        Kalam & \href{https://fonts.google.com}{Google Fonts}  \\
        Lohit Devanagari & \href{https://www.typingguru.net/}{Link}  \\
        Mukta  & \href{https://fonts.google.com}{Google Fonts} \\
        Noto Sans Devanagari & \href{https://fonts.google.com}{Google Fonts} \\
        Palanquin Dark & \href{https://fonts.google.com}{Google Fonts} \\
        Siddhanta & \href{https://www.typingguru.net/}{Link} \\
        Tiro Devanagari Hindi & \href{https://fonts.google.com}{Google Fonts}  \\
    \end{tabular}
    \caption{Fonts used to generate the synthetic dataset (first stage). These fonts were selected to reflect a range of Devanagari styles.}
    \label{tab:fonts}
\end{table}

\textbf{Augmentations used for adding noise to synthetic data generation}
\begin{table}[H]
    \centering
    \begin{tabular}{l|l}
    \textbf{Augmentation} & \textbf{Parameters} \\
    \midrule
    \midrule
    Piecewise Affine & $p = 0.6$, scale = (0.005, 0.015) \\
    Elastic Transformation  & $p = 0.5$, alpha = (1.5, 3.0), sigma = (0.6, 1.0) \\
    Motion Blur    & $p = 0.4$, kernel size $k$ = [3, 5] \\
    Dropout            & $p = 0.4$, dropout rate = (0.01, 0.03) \\
    Gaussian Blur        & $p = 0.4$, sigma = (0.5, 0.9) \\
    Linear Contrast         & $p = 0.3$, factor = (0.7, 1.4) \\
    Brightness Scaling      & $p = 0.3$, multiply factor = (0.85, 1.15) \\
    Laplace Noise        & $p = 0.3$, scale = 0.01 \\
    Vertical Translation    & $p = 0.5$, shift = ±1.5\% \\
    Convolution Blur & 3×3 smoothing kernel \\
    \end{tabular}
    \caption{Augmentation operations used during synthetic dataset generation. $p$ refers to the probability of applying each transformation. Each operation is applied probabilistically using modules from the \texttt{imgaug} library.}
    \label{tab:noise_synth}
\end{table}

\clearpage  
\textbf{Text sources for generating Old Nepali synthetic dataset}

\begin{table}[H]
    \centering
    \resizebox{\columnwidth}{!}{
    \begin{tabular}{l|l}
    \textbf{Name} & \textbf{Source} \\
    \midrule
    \midrule
    Aatma Gyan & \url{https://archive.org/details/aatma-gyan/page/n21/mode/2up} \\
    Antya Karma Paddhati in Nepali &  \url{https://archive.org/details/antya-karma-paddhati-pdf}\\
    Bhagwat Gita in Nepali & \url{https://archive.org/details/20220411_20220411_0657-bhagavad-gita-in-nepali} \\
    Bhanubhakta Ramayan & \url{https://archive.org/details/nepali-bhanubhakta-ramayan} \\
    Chanakya Niti in Nepali & \url{https://archive.org/details/chanakya-niti-in-nepali_hinduism} \\
    Devmala Vamshavali ( Yogi Naraharinath) & \url{https://archive.org/details/DevmalaVamshavaliYogiNaraharinath}\\
    Ekadeshma & \url{https://archive.org/details/ekadeshma_202204}\\
    Gorkha Vamshavali & \url{https://archive.org/details/gorkha-vamshavali}\\
    Grhapravesa & \url{https://archive.org/details/grhapravesa-aitihasika-upanyasa}\\
    Laxmi Prasad Devkota 2025 BS Shakuntala Mahakavya & \url{https://archive.org/details/laxmi-prasad-devkota-2025-bs-shakuntala-mahakavya}\\
    Madhyacandrik & \url{https://archive.org/details/madhyacandrik00sharuoft/page/n5/mode/2up} \\
    Mahabharat in Nepali & \url{https://archive.org/details/mahabharat-sampoorna-18-parva-nepali}\\
    Maharani Rajyalakshmi & \url{https://archive.org/details/kotparva-ki-maharani-rajyalaxmi}\\
    Muna Madan & \url{https://archive.org/details/MunaMadan}\\
    Naridharma tatha Purushartha & \url{https://archive.org/details/20220206_20220206_1126}\\
    Nepal Ritupau 1956 & \url{https://archive.org/details/nepal-ritupau-1956_202204}\\
    Nepal ko itihas & \url{https://archive.org/details/428977-nepal-ko-itihas-1950}\\
    Prameshwar Adalat & \url{https://archive.org/details/Prameshwor-ko-adalat-ma-veda-ra-bible}\\
    Rig Ved in Nepali Language & \url{https://archive.org/details/rig-ved-in-nepali_202112}\\
    Shreemad Bhagwat Mahapuran in nepali & \url{https://archive.org/details/shreemad-bhagwat-mahapuran}\\
    Teachings of Queen Kunti & \url{https://archive.org/details/20220412_20220412_1153-teachings-of-queen-kunti}\\
    \end{tabular}
    }
    \caption{Text sources for generating Old Nepali synthetic corpus and synthetic images.}
    \label{tab:oldNepaliSynth_books}
\end{table}

\begin{figure}[H]
    \centering
    \setlength{\fboxsep}{0pt} 
    \setlength{\fboxrule}{0.5pt} 

    \begin{subfigure}{0.85\textwidth}
        \centering
        \fbox{\includegraphics[width=\linewidth]{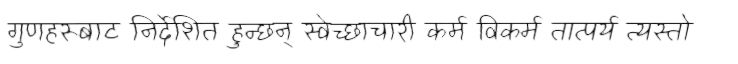}}
        \caption{Sample 1}
    \end{subfigure}

    \vspace{1.5em}

    \begin{subfigure}{0.85\textwidth}
        \centering
        \fbox{\includegraphics[width=\linewidth]{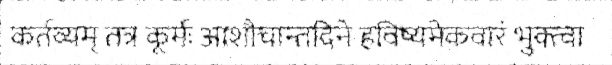}}
        \caption{Sample 2}
    \end{subfigure}

    \vspace{1.5em}

    \begin{subfigure}{0.85\textwidth}
        \centering
        \fbox{\includegraphics[width=\linewidth]{fig_tables/oldNepaliSynth_sample_1.png}}
        \caption{Sample 3}
    \end{subfigure}

    \caption{Three samples of synthetic Devanagari line images generated from the text corpus, which is used in the first stage of training.}

    \label{fig:oldNepaliSynth_samples}
\end{figure}

\clearpage  




\section{Decoding Methods and Results}
\label{a:decoding_results}
\begin{table}[H]
    \centering
    \begin{tabular}{l|l|c|c}
        \textbf{Method} & \textbf{Hyperparameters} & \textbf{Weighted CER} & \textbf{Mean CER} \\ \hline \hline
        Beam Search & \textit{width} = 1 & 0.0483 & 0.0577 \\
        Beam Search & \textit{width} = 5 & 0.0483 & 0.0577 \\
        Beam Search & \textit{width} = 10 & 0.0483 & 0.0577 \\
        Beam Search & \textit{width} = 20 & 0.0483 & 0.0577 \\
        Contrastive Search & $k$ = 5, $\alpha$ = 0.2 & 0.0488 & 0.0581 \\
        Contrastive Search & $k$ = 5, $\alpha$ = 0.6 & 0.0488 & 0.0581 \\
        Contrastive Search & $k$ = 5, $\alpha$ = 0.8 & 0.0488 & 0.0581 \\
        Contrastive Search & $k$ = 10, $\alpha$ = 0.2 & 0.0488 & 0.0581 \\
        Contrastive Search & $k$ = 10, $\alpha$ = 0.6 & 0.0488 & 0.0581 \\
        Contrastive Search & $k$ = 10, $\alpha$ = 0.8 & 0.0488 & 0.0581 \\
        Temperature Sampling & $\tau$ = 0.2 & 0.0488 & 0.0581 \\
        Temperature Sampling & $\tau$ = 0.4 & 0.0488 & 0.0581 \\
        Temperature Sampling & $\tau$ = 0.6 & 0.0488 & 0.0581 \\
        Temperature Sampling & $\tau$ = 0.8 & 0.0488 & 0.0581 \\
        Temperature Sampling & $\tau$ = 0.9 & 0.0485 & 0.0579 \\
        Temperature Sampling & $\tau$ = 1.0 & 0.0488 & 0.0580 \\
        Top-k Sampling & $k$ = 3 & 0.0488 & 0.0571 \\
        Top-k Sampling & $k$ = 5 & 0.0487 & 0.0580 \\
        Top-k Sampling & $k$ = 10 & 0.0489 & 0.0582 \\
        Top-k Sampling & $k$ = 20 & 0.0486 & 0.0579 \\
        Top-k Sampling & $k$ = 50 & 0.0490 &  0.0582 \\
        Top-p Sampling & $p$ = 0.5 & 0.0488 & 0.0580 \\
        Top-p Sampling & $p$ = 0.6 & 0.0488 & 0.0580 \\
        Top-p Sampling & $p$ = 0.7 & 0.0488 & 0.0580 \\
        Top-p Sampling & $p$ = 0.8 & 0.0488 & 0.0580 \\
        Top-p Sampling & $p$ = 0.9 & 0.0488 & 0.0581 \\
        Top-p Sampling & $p$ = 0.95 & 0.0488 & 0.0580 \\
    \end{tabular}
    \caption{Comparison of decoding methods and hyperparameter configurations. All results are reported using the standard model setup with a \texttt{trocr-large-handwritten} encoder, byteBPE tokenizer, and script-aware BERT decoder. Decoding strategies and hyperparameters are based on sensitivity analysis by \citet{garces-arias-etal-2025-decoding}.}
    \label{tab:decoding_full_results}
\end{table}

\clearpage

\section{Benchmarking Analysis}
\label{a:benchmarking}

\textbf{Input sample segmented from Figure~\ref{fig:sample_1}.}
\begin{figure}[H]
    \centering
    \begin{subfigure}[t]{0.9\textwidth}
        \centering
        \setlength{\fboxsep}{0pt} 
        \setlength{\fboxrule}{0.5pt} 
        \fbox{\includegraphics[width=\textwidth]{fig_tables/DNA_sample_2.jpeg}}
        \includegraphics[width=0.8\textwidth]{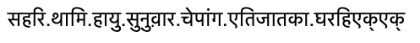}
        \caption{Input sample of manuscript line (top) with ground truth (bottom).}
    \end{subfigure}

    \vspace{1.5em}
    
    \begin{subfigure}[t]{0.9\textwidth}
        \centering
        \includegraphics[width=\textwidth]{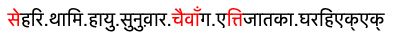}
        \caption{Prediction from fine-tuned TrOCR baseline model.}
    \end{subfigure}
    
    \vspace{1.5em}
    
    \begin{subfigure}[t]{0.9\textwidth}
        \centering
        \includegraphics[width=\textwidth]{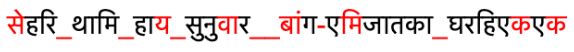}
        \caption{Prediction from Google Cloud Vision API.}
    \end{subfigure}

    \caption{Benchmarking examples showing input and outputs from two OCR baselines. The letters highlighted in red are incorrect predictions.}
    \label{fig:benchmark_examples}
\end{figure}

\section{Evaluating the effect of each training stage}
\label{a:stage_training_effect}
\begin{table}[H]
    \centering
    \begin{tabular}{l|l|l|l}
        \textbf{Training stage} & \textbf{Pretraining} &  \textbf{CER} & \textbf{ACC} \\ \hline \hline
        Stage 1 &  - & 0.71 & 0.0\%  \\
        Stage 2 & Stage 1 & 0.51 & 2.58\%  \\
        \textbf{Stage 3} & \textbf{Stage 1 + Stage 2} & \textbf{0.056} & \textbf{29.58\%}  \\
    \end{tabular}
    \caption{Effect of each training stage on model performance (CER and accuracy), showing how performance improves from Stage 1 to Stage 3 when evaluated on the final (Stage 3) test set.}
    \label{tab:stage_effects}
\end{table}

\clearpage

\section{Error Analysis}
\label{a:error_analysis}

\begin{figure}[H]
    \centering    \includegraphics[width=0.9\textwidth]{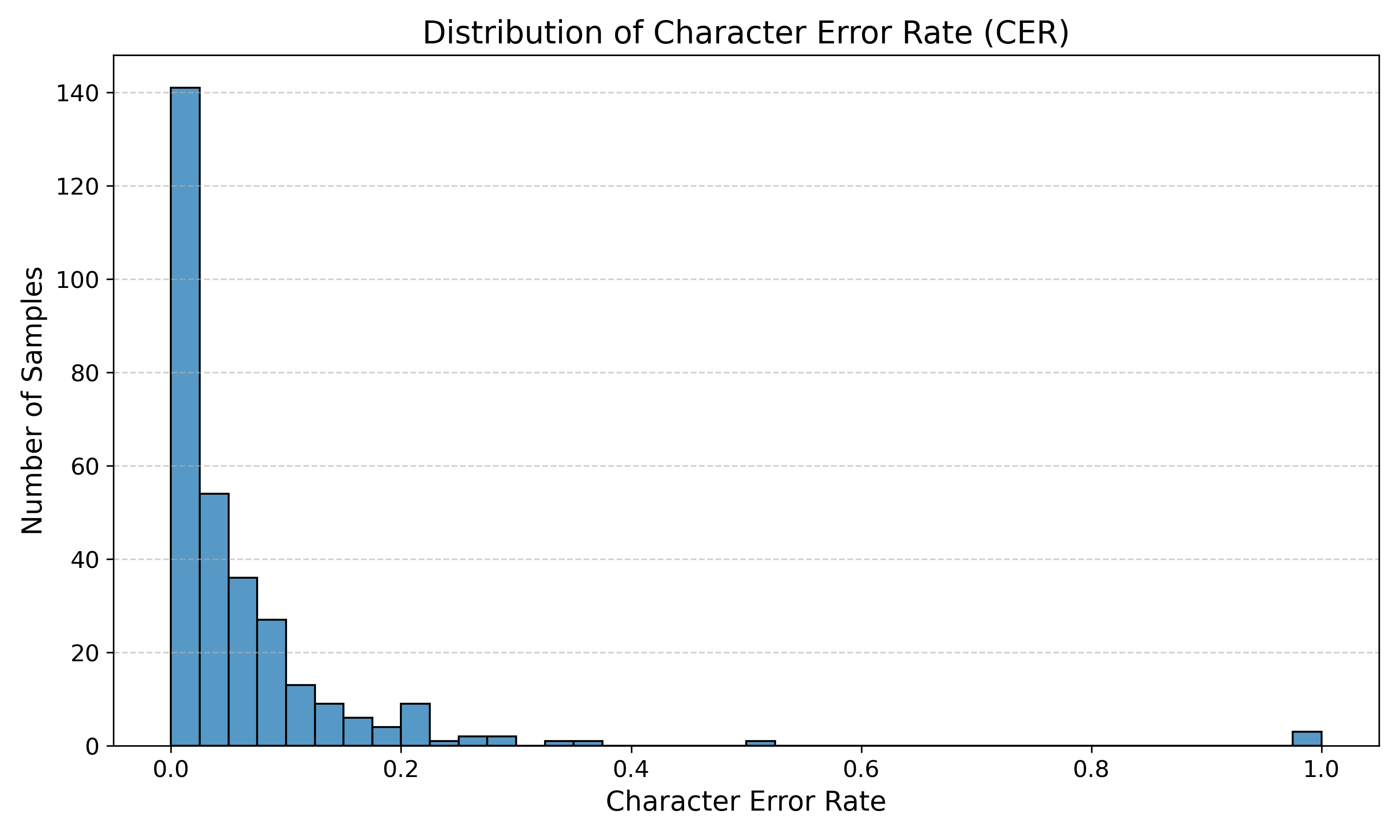}
    \caption{Distribution of CER for all samples in the test set.}
    \label{fig:cer_histogram}
\end{figure}

\begin{figure}[H]
    \centering
    \includegraphics[width=0.9\textwidth]{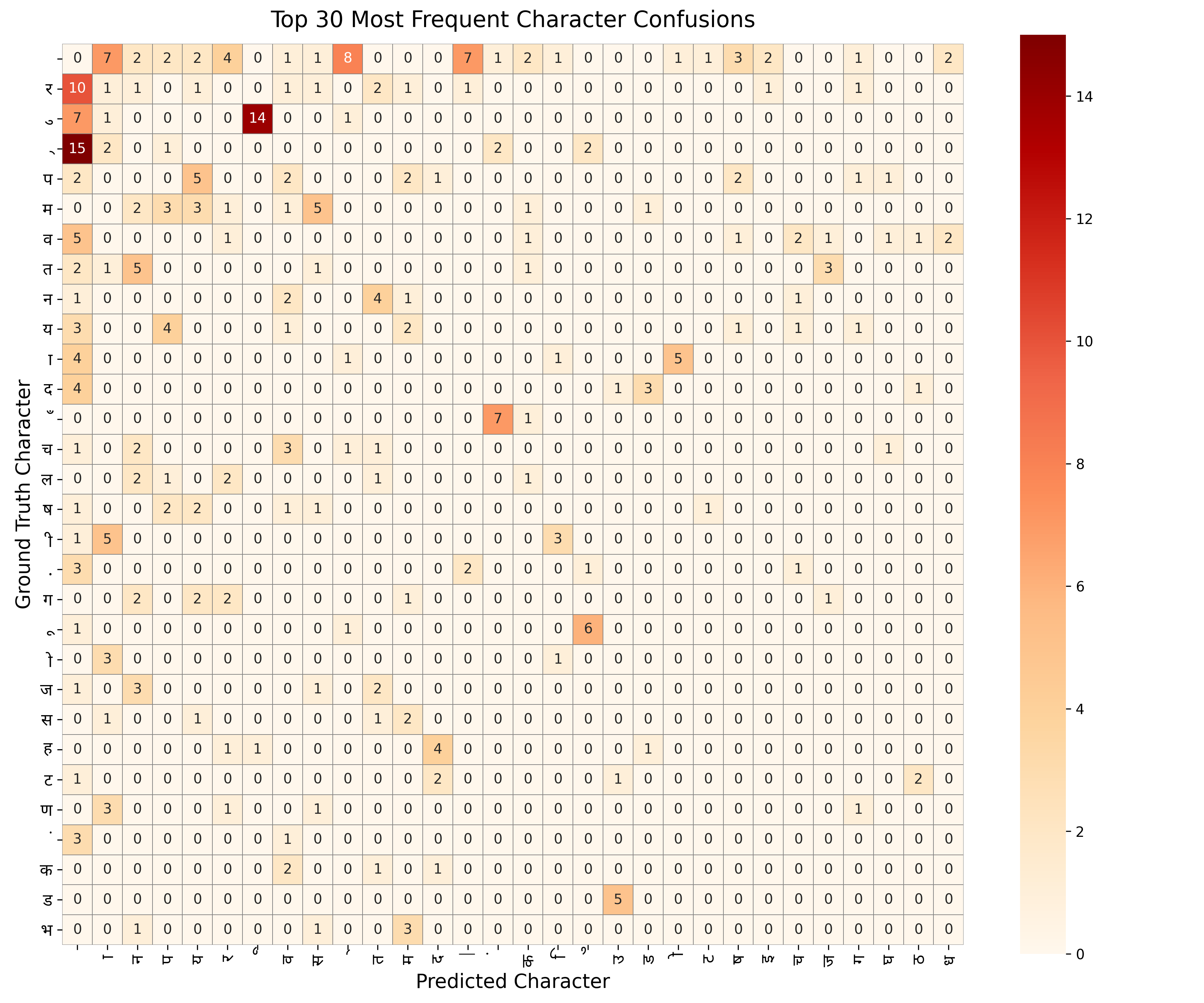}
    \caption{Heatmap showing top 30 most frequent character-level confusions between ground truth and predictions.}
    \label{fig:heatmap}
\end{figure}

\begin{figure}[H]
    \centering
    \includegraphics[width=0.9\textwidth]{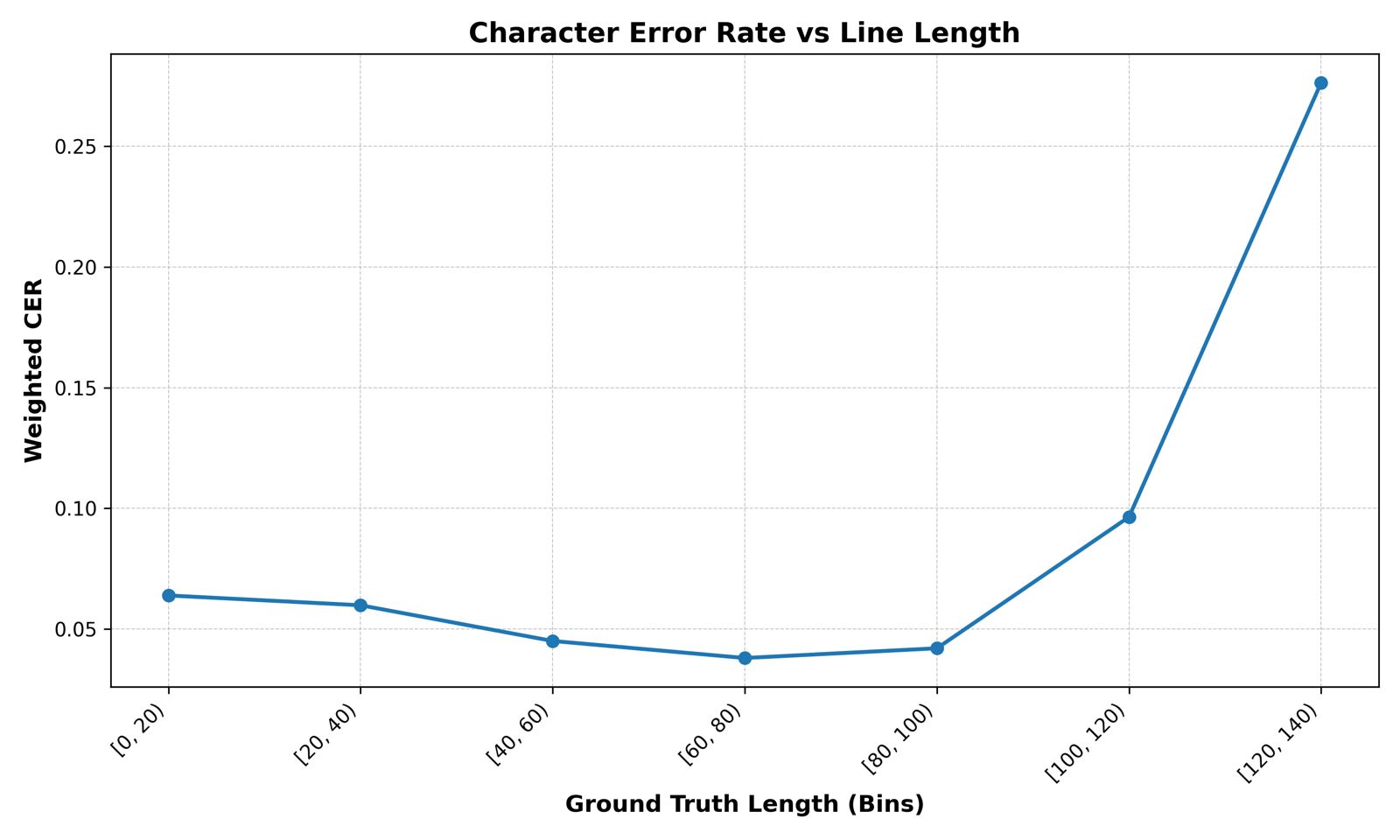}
    \caption{Evaluation of CER against the total number of characters in a line. Each bin groups lines by character count, and the CER is averaged using line lengths as weights.}
    \label{fig:cer_vs_line_length}
\end{figure}

\clearpage

\textbf{Most common errors}
\begin{figure}[H]
    \centering
    \includegraphics[width=0.9\textwidth]{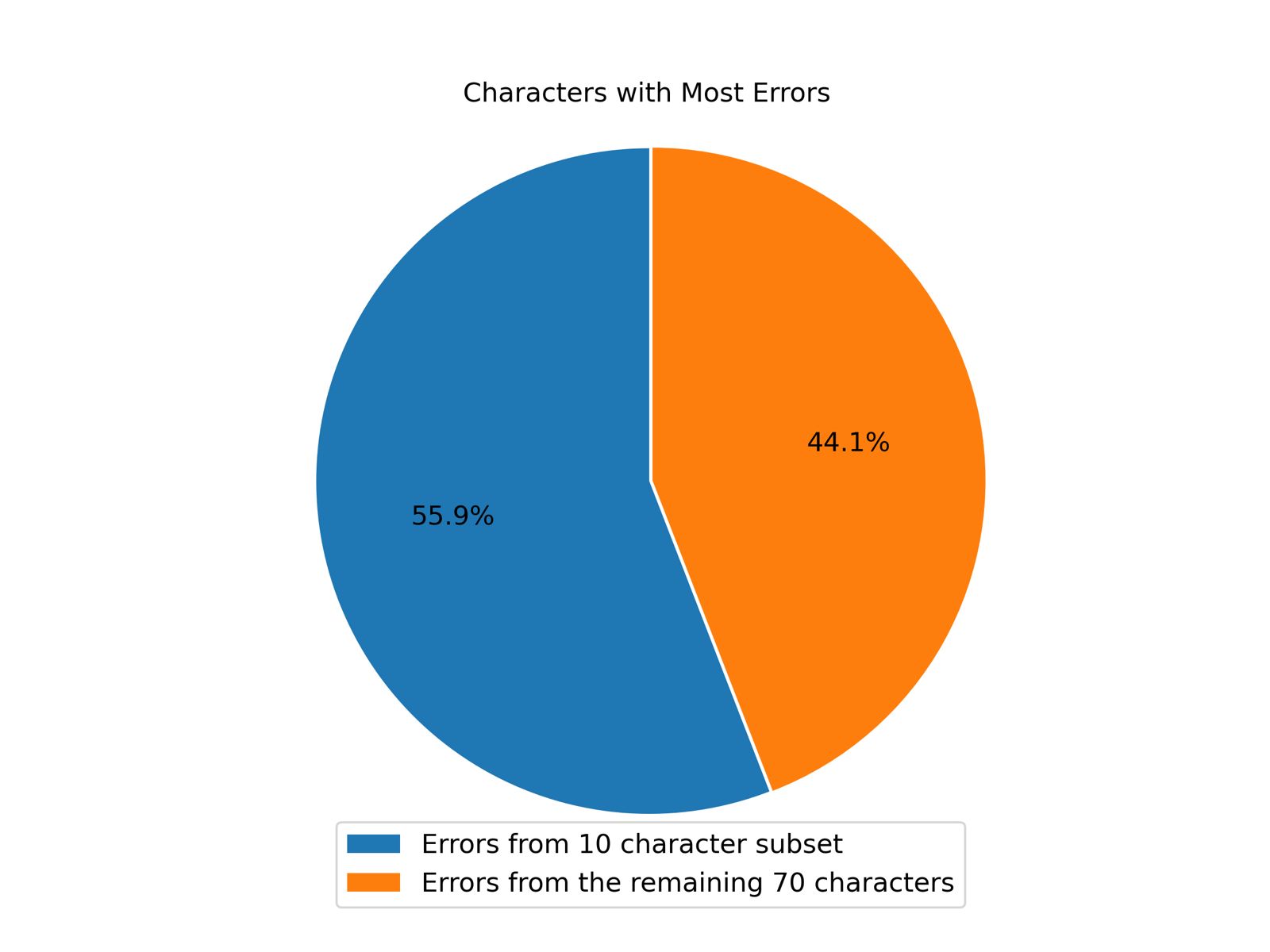}
    \caption{Error share between the top-10 characters with most errors and the remaining 70 characters. The error share is calculated as the ratio of total errors by characters to the total errors in the test set.}
    \label{fig:most_prob_characters}
\end{figure}

\begin{table}[H]
\centering
    \begin{tabular}{l|l|l|l}
              \textbf{Character} & \textbf{Name} & \textbf{Error count} & \textbf{Error share}  \\
              \hline \hline
              \includegraphics[height=10pt]{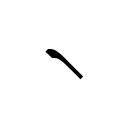} & VIRAMA & 111 & 12.92\% \\
                & SPACE & 98 & 11.41\%  \\
              \includegraphics[height=10pt]{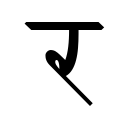} & CONSONANT RA & 56 & 6.52\% \\

              \includegraphics[height=10pt]{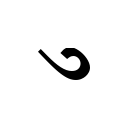} & VOWEL U & 35 & 4.07\% \\
              \includegraphics[height=10pt]{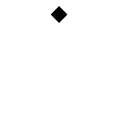} & SIRBINDU & 34 & 3.96\% \\
              \includegraphics[height=10pt]{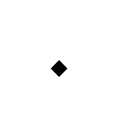} & NUKTA & 33 & 3.84\% \\
              . & FULL STOP & 33 & 3.84\% \\
              \includegraphics[height=10pt]{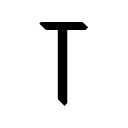} & VOWEL AA &  33 & 3.84\% \\
              \includegraphics[height=10pt]{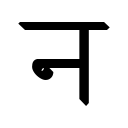} & CONSONANT NA &  25 & 2.79\% \\
                \includegraphics[height=10pt]{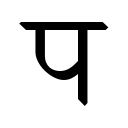} & CONSONANT PA &  23 & 2.68\% \\
              \hline
    \end{tabular}
    \caption{Overview of most common errors with the count of errors in the test set, along with its error share. The total error count is 859.}
    \label{fig:most_prob_characters_table}
\end{table}

\clearpage

\section{Token Uncertainty Analysis}
\label{a:token_uncert_analysis}
\noindent  We analyze whether the model's own uncertainty can be used to flag incorrect predictions. To do this, we compute the relative probability of each predicted token as:

\[
\text{Relative Probability} = \frac{\text{prob}_2}{\text{prob}_1}
\]

\noindent Here, $\text{prob}_1$ and $\text{prob}_2$ refer to the probabilities of the top-1 and top-2 predicted tokens, respectively. These probabilities are extracted from the decoder's output logits, to which we then apply a softmax. We obtain the probability distribution over the vocabulary and then record the probabilities. A high relative probability indicates that the model is uncertain, as it suggests possible confusion between the top two predictions. We therefore assess whether this metric can detect incorrect tokens. We evaluate the performance of this score in distinguishing correct from incorrect tokens using precision, recall, and F1 score. The best F1 score is achieved at a threshold of 0.034 (Table~\ref{fig:token_error_f1}). This threshold is then used to flag low-confidence predictions in further analysis.

\begin{table}[H]
\centering
\begin{tabular}{l|l}
              \textbf{Metric} & \textbf{Score}  \\
              \hline \hline
              F1-score & 0.306 \\
              Precision & 0.363 \\
              Recall & 0.264 \\
\end{tabular}
    \caption{Metrics for detecting incorrect tokens using the model's relative probability between top predictions.}
    
    \label{fig:token_error_f1}
\end{table}

\begin{figure}[H]
    \centering
    \includegraphics[width=0.85\textwidth]{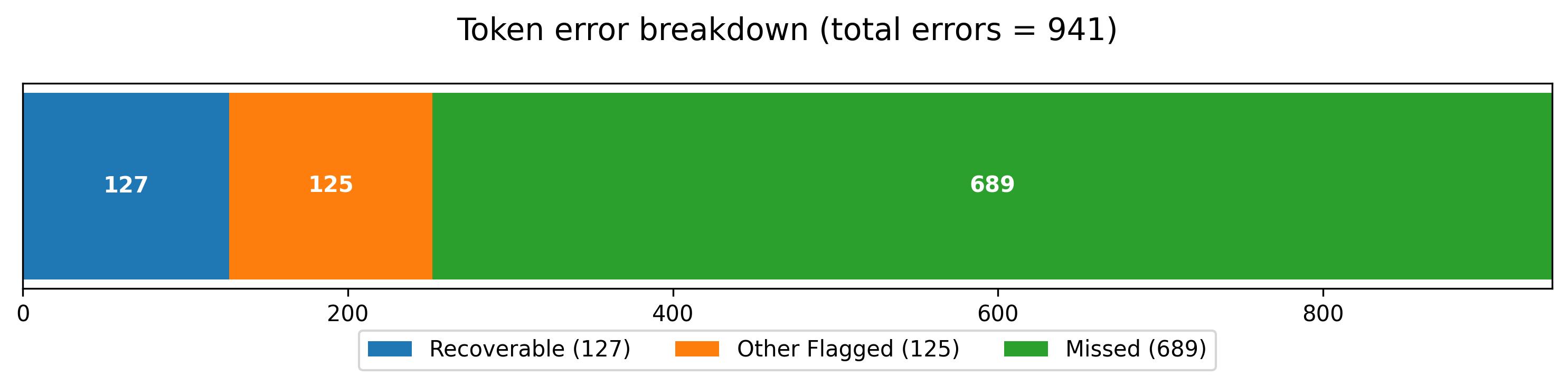}
    \caption{Breakdown of 941 token-level errors based on relative probability. Out of these, 236 tokens were flagged by the model as low-confidence, and 127 of them were recoverable, i.e., the correct token appeared in the top-3 predictions, making them useful for post-correction  or human review. 
    }
    \label{fig:token_error_breakdown}
\end{figure}

\section{Performance Gains}
\label{sec:perf_gains}

\begin{figure}[H]
    \centering
    \includegraphics[width=0.95\textwidth]{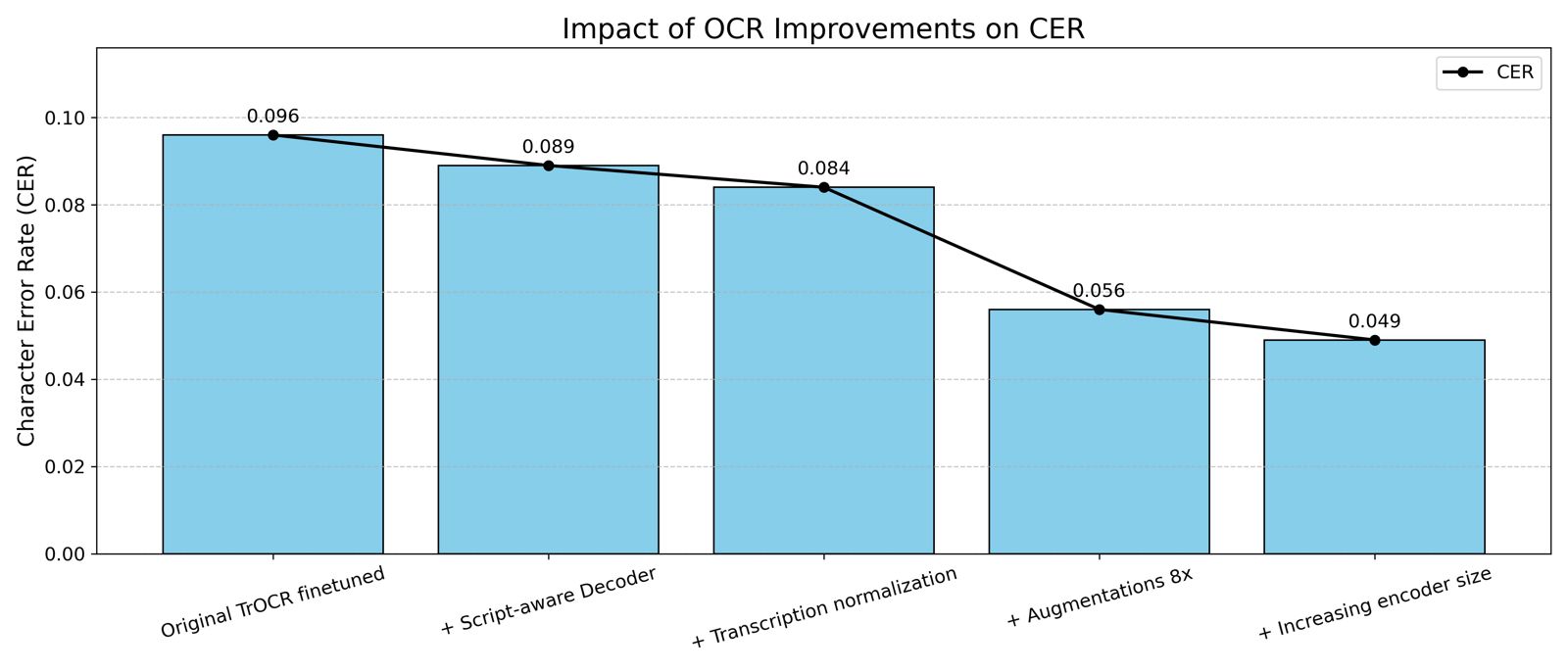}
    \caption{Stepwise reduction in CER through successive improvements to the HTR pipeline.}
    \label{fig:perf_gains}
\end{figure}

\clearpage

\section{Manuscript Samples}
\begin{figure}[H]
    \centering
    \setlength{\fboxsep}{0pt} 
    \setlength{\fboxrule}{0.5pt} 
    \fbox{
    \includegraphics[width=0.75\textwidth]{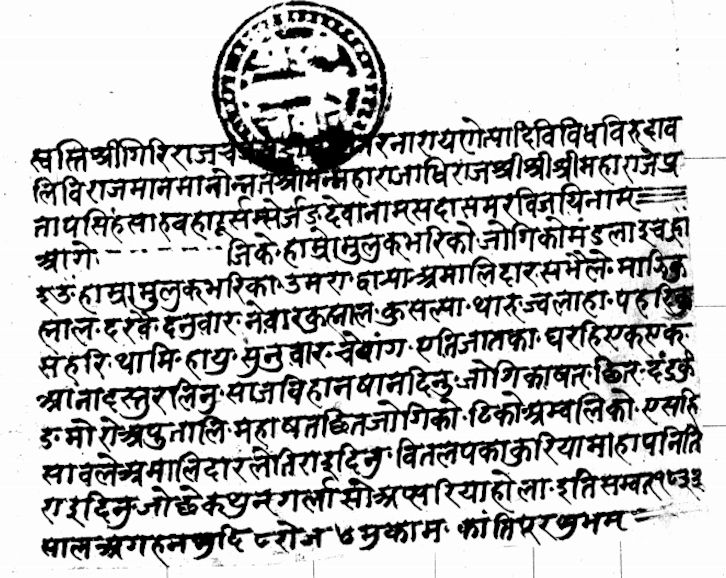}}
    \caption{Manuscript image of NGMPP DNA 14/50 ($\copyright$ National Archives Nepal) excerpted and cropped from \citep{Zotter_2018}. Lines from this paragraph were segmented and used in our OCR dataset, with 80\% allocated to training, 10\% to validation, and 10\% to testing. This figure illustrates a representative sample of our dataset.}
    \label{fig:sample_1}
\end{figure}

\begin{figure}[h]
    \centering
    \setlength{\fboxsep}{0pt} 
    \setlength{\fboxrule}{0.5pt} 
    \fbox{
    \includegraphics[width=0.75\textwidth]{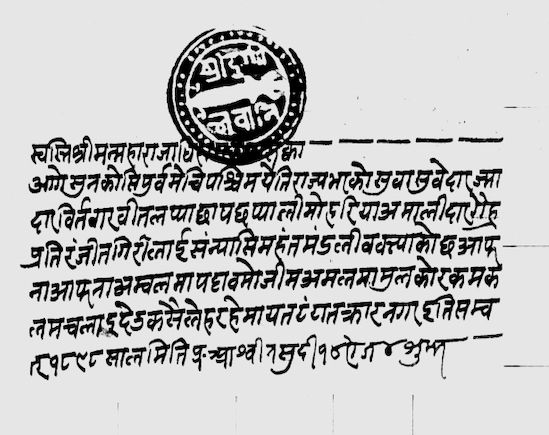}}
    \caption{Another sample of a manuscript image (NGMPP DNA 13/59 $\copyright$ National Archives Nepal) cropped from \citep{Zotter_2018}, also included in our training dataset.}
    \label{fig:sample_2}
\end{figure}
\clearpage
\begin{figure}[h]
    \centering
    \includegraphics[width=0.8\textwidth]{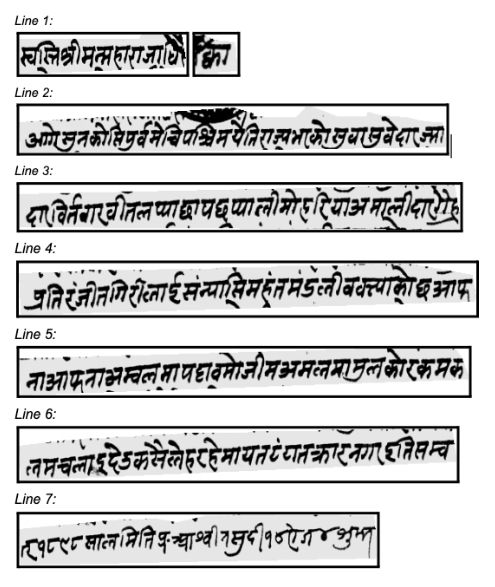}
    \caption{Line-level segmentation of Figure~\ref{fig:sample_2}, performed using Kraken's polygon based method. The text region is segmented into 7 lines, which are used as training samples.}
    \label{fig:cer_vs_length}
\end{figure}

\textbf{Input sample segmented from Figure~\ref{fig:sample_1}}
\begin{figure}[h]
    \centering
    \setlength{\fboxsep}{0pt} 
    \setlength{\fboxrule}{0.5pt} 
    \fbox{
    \includegraphics[width=0.6\textwidth]{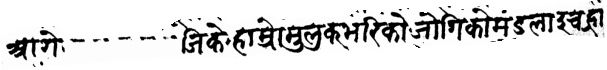}}
    \caption{This image displays the upper dashed line ambiguity, which is normalized to reduce stylistic noise. }
    \label{fig:upper_dash_example}
\end{figure}

\clearpage

\section{Interactive App}
\label{sec:app}

\textbf{Interactive segmentation using Kraken's polygon method}
\begin{figure}[H]
    \centering    
    \setlength{\fboxsep}{0pt} 
    \setlength{\fboxrule}{0.2pt} 
    \fbox{\includegraphics[width=\textwidth]{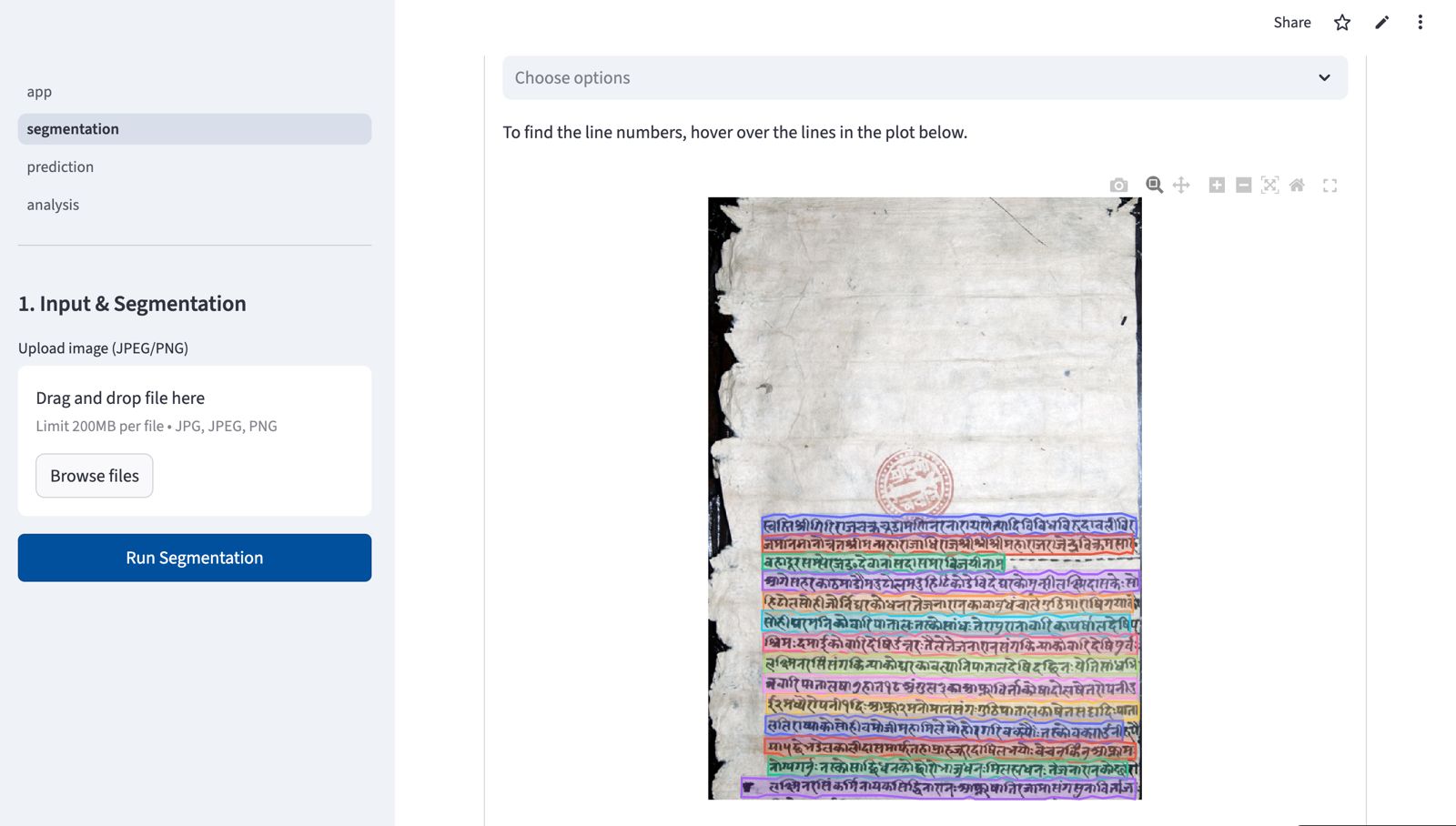}}
    \label{fig:app_1}
\end{figure}

\textbf{Line-by-line HTR prediction on the segmented lines}
\begin{figure}[H]
    \centering    
    \setlength{\fboxsep}{0pt} 
    \setlength{\fboxrule}{0.2pt} 
    \fbox{\includegraphics[width=\textwidth]{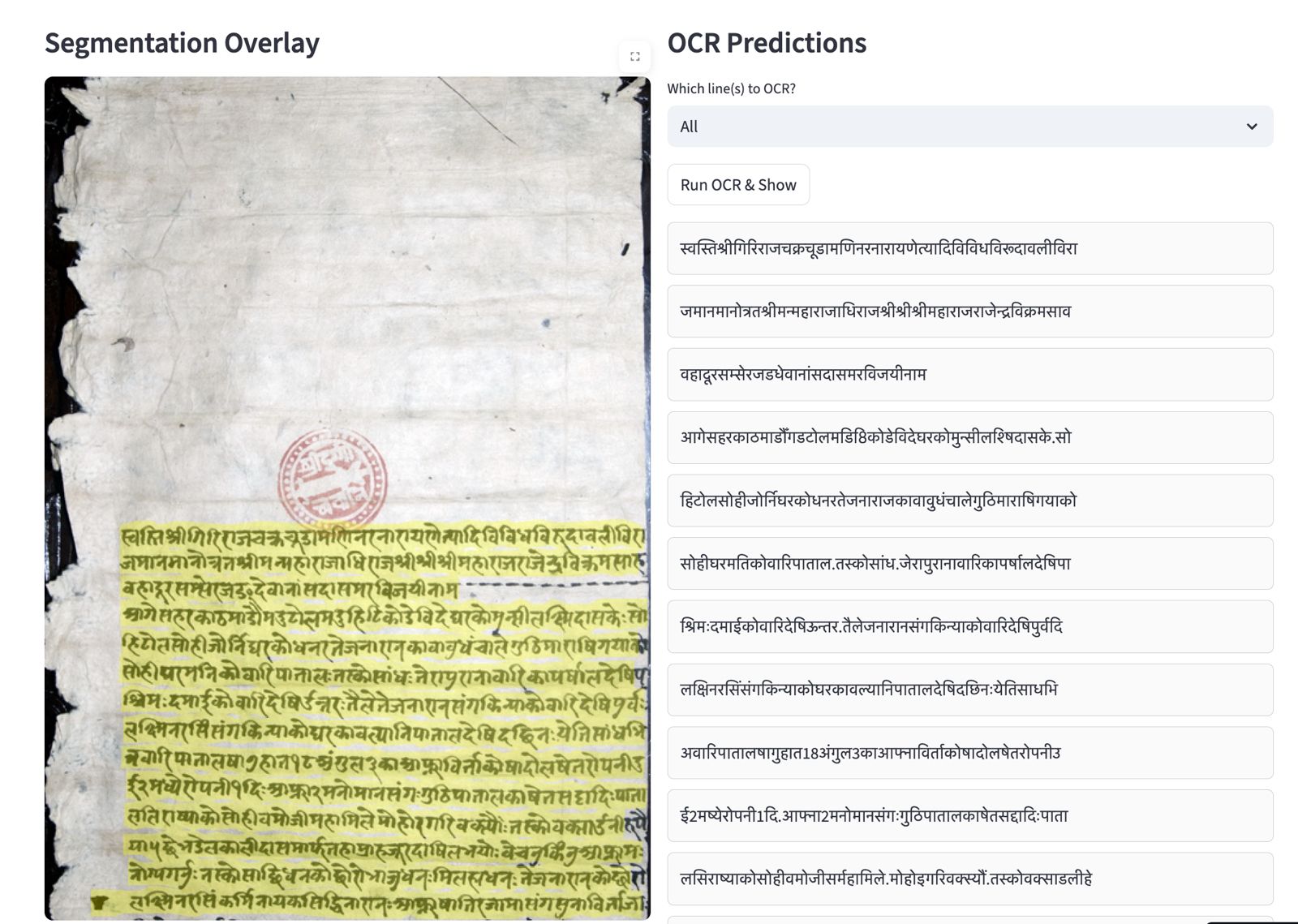}}
    \label{fig:app_2}
\end{figure}

\end{document}

%% file: fig_tables/data_config.tex
\begin{tabular}{|c|c|c|c|}
\cline{2-4}
\multicolumn{1}{c|}{} & \textbf{Train} & \textbf{Eval} & \textbf{Test} \\
\hline
\textbf{First Stage: Pre-training} & 100,000 & 2500 & 2500 \\
\multicolumn{1}{|c|}{\small (Synthetic Devanagari images)} & & & \\
\hline
\textbf{Second Stage: Transfer Learning} & 4,111 & 514 & 514 \\
\multicolumn{1}{|c|}{\small \qquad (Printed Nagari scripts)} & & & \\
\hline
\textbf{Third Stage: Final Model Training} & 2,480 & 310  & 310 \\
\multicolumn{1}{|c|}{\small (Handwritten Old Nepali images)} & & & \\
\hline
\textbf{Total} & \textbf{106,591} & \textbf{3,324} & \textbf{3,324} \\
\hline
\end{tabular}

%% file: fig_tables/results_merged_valid.tex
\begin{tabular}{|c|c|c|c|c|c|c|c|c|}
\hline
\textbf{Encoder} 
  & \textbf{Decoder} & \textbf{Tokenizer} & \textbf{Training stage} 
  & \textbf{Train time (hrs)} & \textbf{Eval runtime (s)} 
  & \textbf{CER} & \textbf{CER(w)} & \textbf{ACC} \\
\hline
\multirow{12}{*}{\texttt{trocr-base-handwritten}}
  & \multirow{6}{*}{BERT}
  & \multirow{3}{*}{byteBPE}
    & Pretraining w/ synthetic data & 2.45 & 344.77 & 0.005 & 0.006 & 90.6\%\\
  &                       & 
    & Transfer Learning w/ printed Nagari scripts & 0.40 & 60.31 & 0.017 & 0.019 & 68.2\% \\
  &                       & 
    & Final model training w/ Old Nepali scripts & 0.69 & 61.66 & \textbf{0.082}  & 0.082 & 24.8\%\\
\cline{3-9}
  &                       & \multirow{3}{*}{charBPE}
    & Pretraining w/ synthetic data & 2.35 & 268.18 & 0.007 & 0.007  & 90.5\%\\
  &                       & 
    & Transfer Learning w/ printed Nagari scripts  & 0.37 & 47.72 & 0.017 & 0.018  &  70.7\% \\
  &                       & 
    & Final model training w/ Old Nepali scripts & 0.56 & 39.88 & 0.087 & 0.086  & 25.5\%\\
\cline{2-9}
  & \multirow{6}{*}{GPT-2}
  & \multirow{3}{*}{byteBPE}
    & Pretraining w/ synthetic data & 2.13 & 282.74 & 0.007  & 0.007 & 90.9\%\\
  &                       & 
    & Transfer Learning w/ printed Nagari scripts  & 0.39 & 51.12 & 0.018 & 0.019  & 66.1\%\\
  &                       & 
    & Final model training w/ Old Nepali scripts & 0.65 & 54.26 &  0.084 & 0.084 & 26.1\%\\
\cline{3-9}
  &                       & \multirow{3}{*}{charBPE}
    & Pretraining w/ synthetic data & 2.34 & 217.95 & 0.007 & 0.009 & 90.7\%\\
  &                       & 
    & Transfer Learning w/ printed Nagari scripts  & 0.36 & 42.41 & 0.017 & 0.017 & 72.7\% \\
  &                       & 
    & Final model training w/ Old Nepali scripts & 0.54 & 34.37 &  0.084 & 0.082 & 28.7\%\\
\hline
\end{tabular}

%% file: fig_tables/char_freq.tex
\begin{tabular}{cccrccccr}
\cline{1-4} \cline{6-9}
\textbf{\#} & \textbf{Character} & \textbf{Frequency} & \textbf{Relative frequency (\%)} &  & \textbf{\#} & \textbf{Character} & \textbf{Frequency} & \textbf{Relative frequency (\%)} \\
\cline{1-4} \cline{6-9}
\noalign{\vskip 1ex}
1 & \includegraphics[width=1em]{char_images/01_.png} & 19516 & 11.5030 & & 2 & \includegraphics[width=1em]{char_images/02_.png} & 13258 & 7.8145 \\
3 & \includegraphics[width=1em]{char_images/03_ra.png} & 11403 & 6.7211 & & 4 & \includegraphics[width=1em]{char_images/04_na.png} & 7352 & 4.3334 \\
5 & \includegraphics[width=1em]{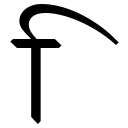} & 7352 & 4.3334 & & 6 & \includegraphics[width=1em]{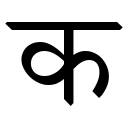} & 6798 & 4.0068 \\
7 & \includegraphics[width=1em]{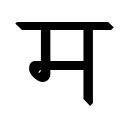} & 5749 & 3.3885 & & 8 & \includegraphics[width=1em]{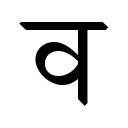} & 5575 & 3.2860 \\
9 & \includegraphics[width=1em]{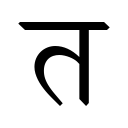} & 5447 & 3.2105 & & 10 & \includegraphics[width=1em]{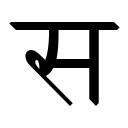} & 5025 & 2.9618 \\
11 & \includegraphics[width=1em]{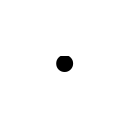} & 4768 & 2.8103 & & 12 & \includegraphics[width=1em]{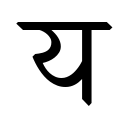} & 4767 & 2.8097 \\
13 & \includegraphics[width=1em]{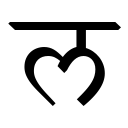} & 4726 & 2.7856 & & 14 & \includegraphics[width=1em]{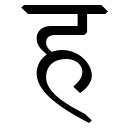} & 4281 & 2.5233 \\
15 & \includegraphics[width=1em]{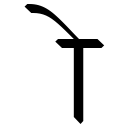} & 4158 & 2.4508 & & 16 & \includegraphics[width=1em]{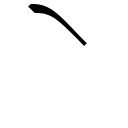} & 4153 & 2.4478 \\
17 & \includegraphics[width=1em]{char_images/17_.png} & 3997 & 2.3559 & & 18 & \includegraphics[width=1em]{char_images/18_pa.png} & 3849 & 2.2687 \\
19 & \includegraphics[width=1em]{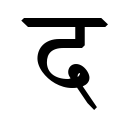} & 3833 & 2.2592 & & 20 & \includegraphics[width=1em]{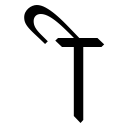} & 3527 & 2.0789 \\
21 & \includegraphics[width=1em]{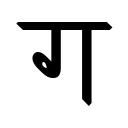} & 3294 & 1.9415 & & 22 & \includegraphics[width=1em]{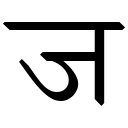} & 3192 & 1.8814 \\
23 & \includegraphics[width=1em]{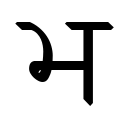} & 2194 & 1.2932 & & 24 & \includegraphics[width=1em]{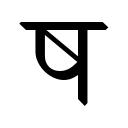} & 1902 & 1.1211 \\
25 & \includegraphics[width=1em]{char_images/25_.png} & 1783 & 1.0509 & & 26 & \includegraphics[width=1em]{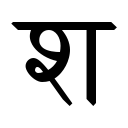} & 1536 & 0.9053 \\
27 & \includegraphics[width=1em]{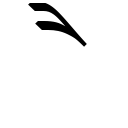} & 1461 & 0.8611 & & 28 & \includegraphics[width=1em]{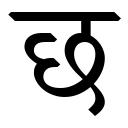} & 1166 & 0.6873 \\
29 & \includegraphics[width=1em]{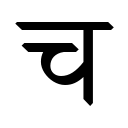} & 1092 & 0.6436 & & 30 & \includegraphics[width=1em]{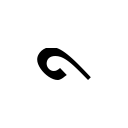} & 966 & 0.5694 \\
31 & \includegraphics[width=1em]{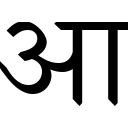} & 933 & 0.5499 & & 32 & \includegraphics[width=1em]{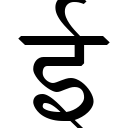} & 905 & 0.5334 \\
33 & \includegraphics[width=1em]{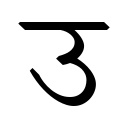} & 886 & 0.5222 & & 34 & \includegraphics[width=1em]{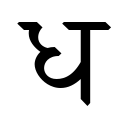} & 868 & 0.5116 \\
35 & \includegraphics[width=1em]{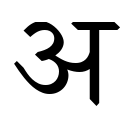} & 867 & 0.5110 & & 36 & \includegraphics[width=1em]{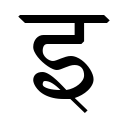} & 799 & 0.4709 \\
37 & \includegraphics[width=1em]{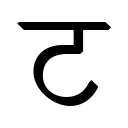} & 796 & 0.4692 & & 38 & \includegraphics[width=1em]{char_images/38_.png} & 699 & 0.4120 \\
39 & \includegraphics[width=1em]{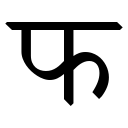} & 691 & 0.4073 & & 40 & \includegraphics[width=1em]{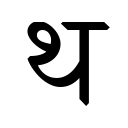} & 678 & 0.3996 \\
41 & \includegraphics[width=1em]{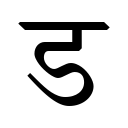} & 673 & 0.3967 & & 42 & \includegraphics[width=1em]{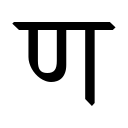} & 643 & 0.3790 \\
43 & \includegraphics[width=1em]{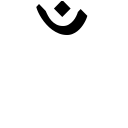} & 569 & 0.3354 & & 44 & \includegraphics[width=1em]{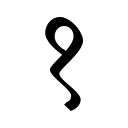} & 562 & 0.3313 \\
45 & \includegraphics[width=1em]{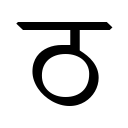} & 491 & 0.2894 & & 46 & \includegraphics[width=1em]{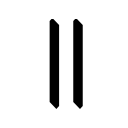} & 426 & 0.2511 \\
47 & \includegraphics[width=1em]{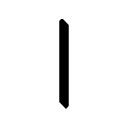} & 376 & 0.2216 & & 48 & \includegraphics[width=1em]{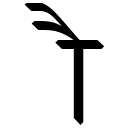} & 310 & 0.1827 \\
49 & \includegraphics[width=1em]{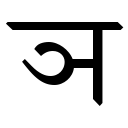} & 306 & 0.1804 & & 50 & \includegraphics[width=1em]{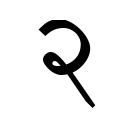} & 275 & 0.1621 \\
51 & \includegraphics[width=1em]{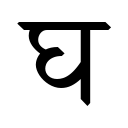} & 259 & 0.1527 & & 52 & \includegraphics[width=1em]{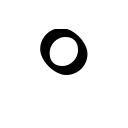} & 198 & 0.1167 \\
53 & \includegraphics[width=1em]{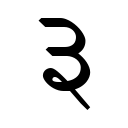} & 192 & 0.1132 & & 54 & \includegraphics[width=1em]{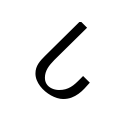} & 187 & 0.1102 \\
55 & \includegraphics[width=1em]{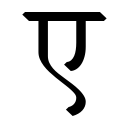} & 183 & 0.1079 & & 56 & \includegraphics[width=1em]{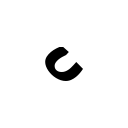} & 177 & 0.1043 \\
57 & \includegraphics[width=1em]{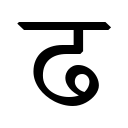} & 176 & 0.1037 & & 58 & \includegraphics[width=1em]{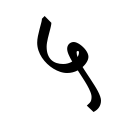} & 174 & 0.1026 \\
59 & \includegraphics[width=1em]{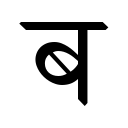} & 147 & 0.0866 & & 60 & \includegraphics[width=1em]{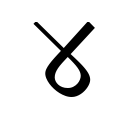} & 136 & 0.0802 \\
61 & \includegraphics[width=1em]{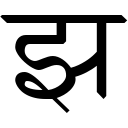} & 124 & 0.0731 & & 62 & \includegraphics[width=1em]{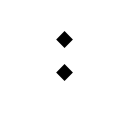} & 123 & 0.0725 \\
63 & \includegraphics[width=1em]{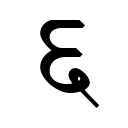} & 121 & 0.0713 & & 64 & \includegraphics[width=1em]{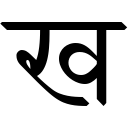} & 116 & 0.0684 \\
65 & \includegraphics[width=1em]{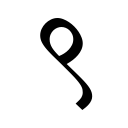} & 112 & 0.0660 & & 66 & \includegraphics[width=1em]{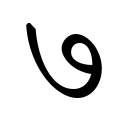} & 101 & 0.0595 \\
67 & \includegraphics[width=1em]{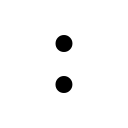} & 82 & 0.0483 & & 68 & \includegraphics[width=1em]{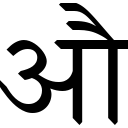} & 69 & 0.0407 \\
69 & \includegraphics[width=1em]{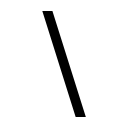} & 43 & 0.0253 & & 70 & \includegraphics[width=1em]{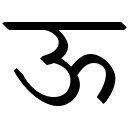} & 37 & 0.0218 \\
71 & \includegraphics[width=1em]{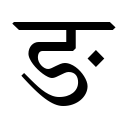} & 35 & 0.0206 & & 72 & \includegraphics[width=1em]{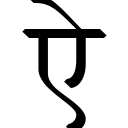} & 24 & 0.0141 \\
73 & \includegraphics[width=1em]{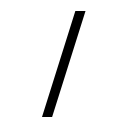} & 18 & 0.0106 & & 74 & \includegraphics[width=1em]{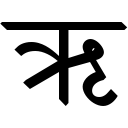} & 13 & 0.0077 \\
75 & \includegraphics[width=1em]{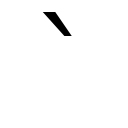} & 13 & 0.0077 & & 76 & \includegraphics[width=1em]{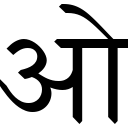} & 7 & 0.0041 \\
77 & \includegraphics[width=1em]{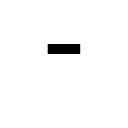} & 5 & 0.0029 & & 78 & \includegraphics[width=1em]{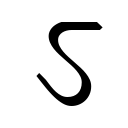} & 4 & 0.0024 \\
79 & \includegraphics[width=1em]{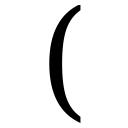} & 1 & 0.0006 & & 80 & \includegraphics[width=1em]{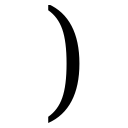} & 1 & 0.0006 \\
\cline{1-4} \cline{6-9}
\end{tabular}